\PassOptionsToPackage{table}{xcolor}
\documentclass[]{TJURLLAB}

\usepackage{algorithm}
\usepackage{algpseudocode}
\usepackage{listings}
\usepackage{wrapfig}
\usepackage{xspace}
\usepackage{tabularx}
\usepackage{array}
\usepackage{nicefrac}
\usepackage{fancyhdr}
\usepackage{fontawesome5}

\fancypagestyle{plain}{\fancyhf{}\fancyfoot[C]{\thepage}}
\definecolor{appendixpurple}{RGB}{116,54,176}
\definecolor{appendixlightpurple}{RGB}{248,244,255}
\definecolor{darkmagenta}{rgb}{0.55,0.0,0.55}
\definecolor{axisbluebg}{RGB}{214,234,248}
\definecolor{axisgreenbg}{RGB}{212,239,223}
\definecolor{axisamberbg}{RGB}{253,235,213}
\newcommand{\axisA}{\colorbox{axisbluebg}{\textbf{(A)}}}
\newcommand{\axisB}{\colorbox{axisgreenbg}{\textbf{(B)}}}
\newcommand{\axisC}{\colorbox{axisamberbg}{\textbf{(C)}}}

\definecolor{Cornsilk}{rgb}{1.0, 0.97, 0.86}
\definecolor{lightorange}{rgb}{0.996, 0.855, 0.643}
\definecolor{lightgray}{rgb}{1.0, 0.827, 0.278}
\newcommand{\method}{\textbf{WarpSAC}\xspace}
\lstdefinestyle{pythonappendix}{
    language=Python,
    basicstyle=\ttfamily\scriptsize,
    keywordstyle=\color{appendixpurple}\bfseries,
    commentstyle=\color{gray},
    stringstyle=\color{darkmagenta},
    showstringspaces=false,
    breaklines=true,
    columns=fullflexible,
    frame=single,
    rulecolor=\color{appendixpurple},
    backgroundcolor=\color{appendixlightpurple},
    xleftmargin=0.5em,
    xrightmargin=0.5em,
    belowskip=0.25em,
    aboveskip=0.25em
}

\newcommand{\appentry}[3]{%
\hyperref[#3]{\textcolor{appendixpurple}{\Large\textbf{#1}}} &
\hyperref[#3]{\textcolor{appendixpurple}{\textbf{#2}}} &
\hyperref[#3]{\textcolor{appendixpurple}{\textbf{\pageref*{#3}}}} \\
}

\newcommand{\appsubentry}[3]{%
&
\hspace{1.4em}\hyperref[#3]{\textcolor{appendixpurple}{#1\quad #2}} &
\hyperref[#3]{\textcolor{appendixpurple}{\pageref*{#3}}} \\
}

\setlogo{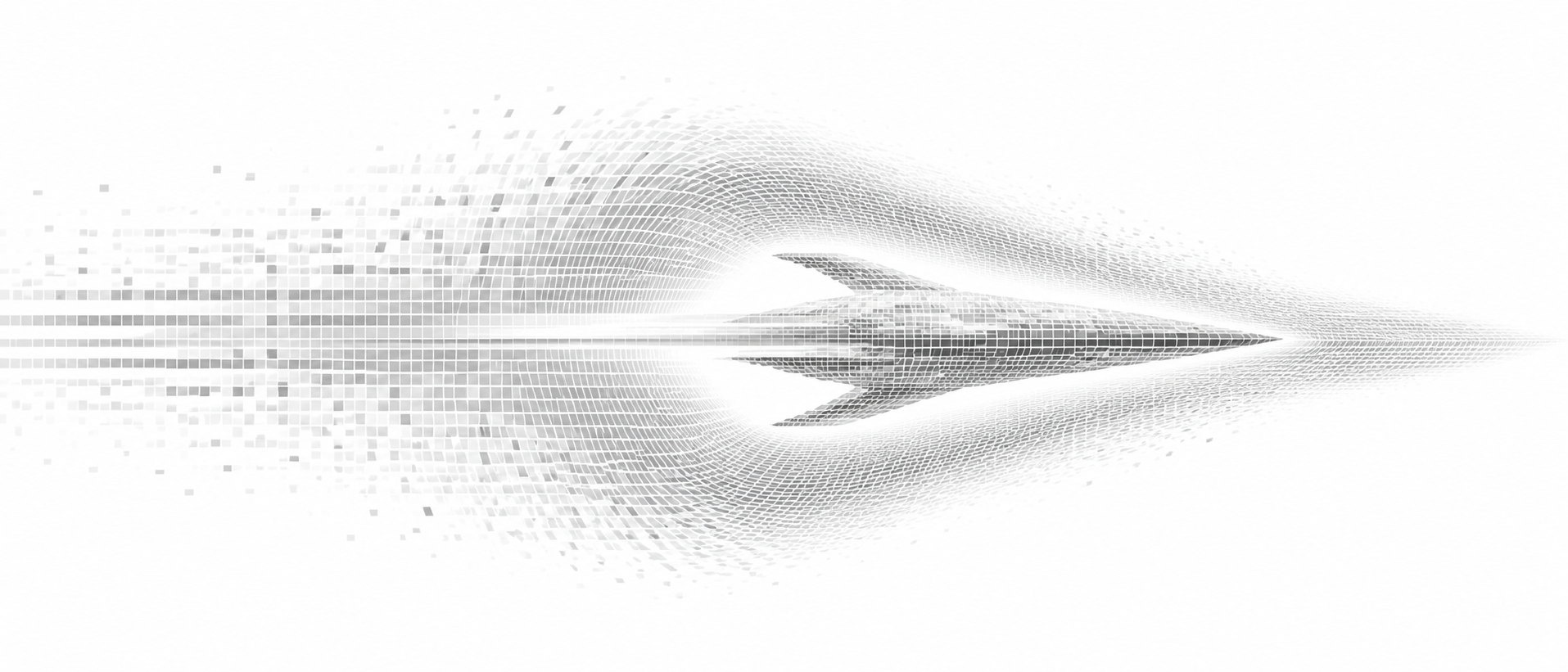}

\title{WarpSAC\footnotemark: Towards the Pinnacle of Scalable Off-policy RL by Rethinking Exploration and Exploitation}

\author[1]{Zihao Wu}
\author[1,\mbox{\scriptsize\faEnvelope}]{Hongyao Tang}
\author[2,\mbox{\scriptsize\faEnvelope}]{Yi Ma}
\author[2]{Huizhong Song}
\author[1]{Pengyi Li}
\author[1]{Yifu Yuan}
\author[3]{Fei Ni}
\author[1]{Jinyi Liu}
\author[2]{Wei Wei}
\author[1]{Jianrong Wang}
\author[1]{Yan Zheng}
\author[1,\mbox{\scriptsize\faEnvelope}]{Jianye Hao}

\affiliation[1]{Tianjin University}
\affiliation[2]{Shanxi University}
\affiliation[3]{Imperial College London}

\contribution[\faEnvelope]{Corresponding Author(s)}

\abstract{\subsubsection*{Abstract}
Scaling off-policy reinforcement learning (RL) through massively parallel simulation changes the data regime assumptions, under which RL algorithms are designed. Canonical stabilizers  are motivated by data-limited training, where replay buffers provide narrow state--action coverage.
By contrast, massively parallel simulation offers diverse experience at high throughput, naturally challenging the canonical roles of the stabilizers in this new data regime.
Through comprehensive and controlled empirical study across eight benchmark families spanning CPU-scale locomotion, GPU-parallel robotic simulation, dexterous manipulation, humanoid whole-body control and we find that these stabilizers are strongly \textit{data-regime-dependent}: parameter normalization helps under narrow replay coverage but restricts value fitting when data are abundant, clipped double-Q can be safely relaxed in high-throughput manipulation, and age-biased replay weighting is broadly useful for improving learning efficiency, especially under limited network capacity.
Turning this analysis into a prescription, we build \method, a regime-aware family of off-policy RL algorithms.
\method{} uses Sample Weight Decay as a regime-agnostic component for efficient exploitation and matches each regime with a prescribed variant: \textbf{\method{}-L} (Norm ON, clipped double-Q) for data-limited CPU-scale training, and \textbf{\method{}-A} (Norm OFF, single-Q) for data-abundant GPU-parallel training.
\textcolor{appendixpurple}{\method{} improves normalized score--step AUC over FlashSAC by $4.5\%$ across \textbf{nine CPU-scale environments} and $23.1\%$ across \textbf{fourteen GPU-parallel environments}; lifts UnitreeG1TransportBox-v1 success rate from $19.8\%$ to $96.4\%$, and gains $19.1\%$ in mean normalized wall-time AUC on \textbf{MuJoCo Playground}; achieves \textbf{faster sim-to-real deployment} on \texttt{Unitree G1} than FlashSAC by 36.4\% in terms of wall time}. These results argue that scalable off-policy RL should adapt its stabilizers to the available data regime. Under this principle, 
\method{} advances the state of the art of scalable off-policy RL, delivering consistent gains over FlashSAC across different data regimes.
}

\correspondence{\href{mailto:wzhhasadream@gmail.com}{\textcolor{accentcolor}{\texttt{wzhhasadream@gmail.com}}}}
\metadata[Code]{\href{https://github.com/wzhhasadream/warprl}{\textcolor{accentcolor}{https://github.com/wzhhasadream/warprl}}}
\metadata[Project Page]{\href{https://wzhhasadream.github.io/WarpSAC/}{\textcolor{accentcolor}{https://wzhhasadream.github.io/WarpSAC/}}}

\begin{document}
\thispagestyle{plain}
\maketitle

\begin{figure}[H]
    \centering
    \vspace{-0.5cm}
    \includegraphics[width=\linewidth]{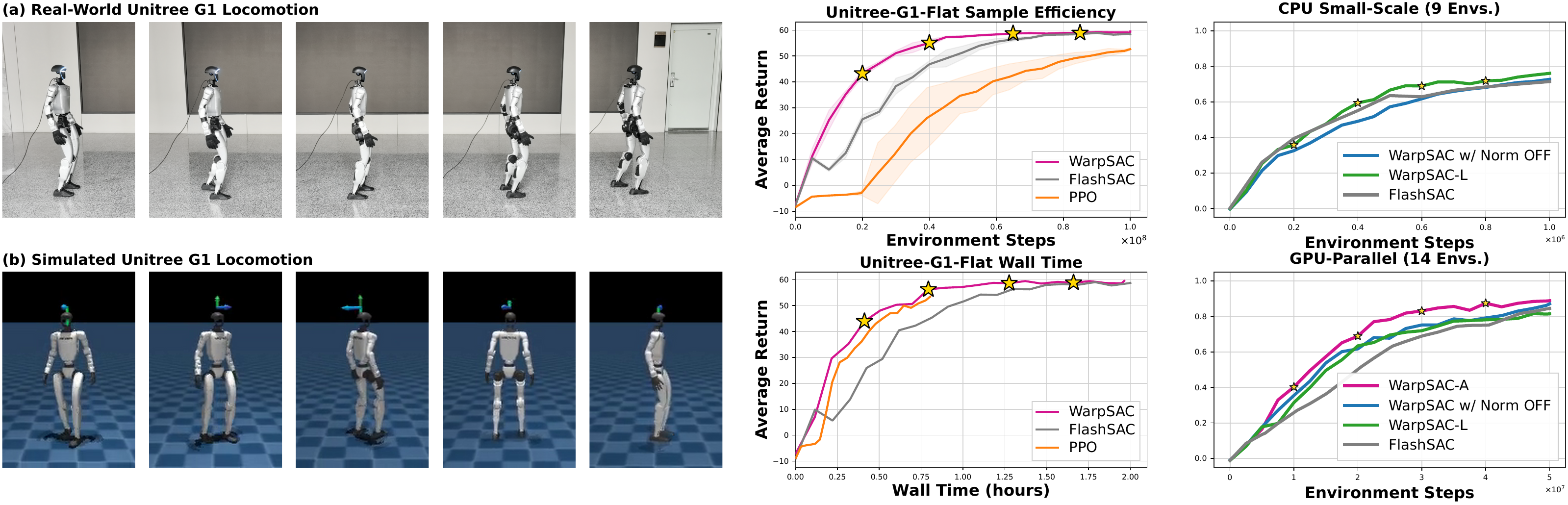}
    \vspace{-0.7cm}
    \caption{\textbf{Benchmark-scale and sim-to-real overview.}
    The left strips show real-world and simulated Unitree G1 locomotion, the middle panels report Unitree-G1-Flat sample efficiency and training wall time measured on a single NVIDIA A800 GPU, and the right panels summarize normalized CPU-scale and GPU-parallel learning curves. Across the benchmark curves, \method improves over FlashSAC, while the preferred stabilizing configuration depends on the data-collection regime.}
    \label{fig:intro}
    \vspace{-0.6cm}
\end{figure}

\footnotetext{``Warp'', i.e., warp drive or warp speed, metaphorically denotes faster scalable off-policy learning, echoing ``Warp Drive, Mr. Scott. Ahead Warp One, Mr. Sulu.'' by Captain James T. Kirk, \textit{Star Trek: The Motion Picture (1979)}.}

\section{Introduction}

Scaling has become a central force in modern reinforcement learning (RL) for robot control. GPU-accelerated simulators and massively parallel environments now let agents collect diverse experience at rates that were unattainable in conventional single-environment training~\citep{mittal2025isaaclab,zakka2025mujoco,zakka2026mjlab,taomaniskill3}. This shift makes off-policy actor--critic methods increasingly attractive: replay buffers can reuse large volumes of interaction data, while scalable implementations such as Soft Actor-Critic (SAC)~\citep{haarnoja2018sac,haarnoja2018applications} and FlashSAC~\citep{kim2026flashsac} have shown that off-policy learning can be effective in high-dimensional robotic domains.

However, most classical off-policy stabilizers were designed under the opposite assumption: that replay coverage is narrow, and the learner must protect itself against extrapolation. In this limited data regime, exploration and conservative value estimation are essential.
Entropy regularization pushes the policy toward unseen actions, clipped double-Q targets suppress overestimation on poorly covered transitions, and parameter normalization or norm control constrains the function class so that value fitting remains stable outside the replay distribution~\citep{fujimoto2018td3,haarnoja2018sac,lee2025hyperspherical}. 
When massively parallel simulation replaces this scarcity with abundance, the same mechanisms may no longer be uniformly beneficial, as the bottleneck shifts from exploration and conservatism to exploiting abundant data efficiently. Strong parameter normalization can restrict expressive freedom; clipped double-Q can be overly pessimistic; and every extra conservative component adds wall-clock cost. Yet almost all scalable off-policy pipelines still inherit these stabilizers unchanged. The central question we ask in this paper is --- \textit{not whether these stabilizers are good in isolation, but when their benefits outweigh their restrictions.}

We answer this question through a controlled component-wise analysis based on FlashSAC~\citep{kim2026flashsac}. We isolate three axes that scalable off-policy learners usually bundle together: (i) replay-side data utilization, instantiated by Sample Weight Decay (SWD)~\citep{wu2026rank}, an age-biased sampler that concentrates updates on policy-relevant transitions; (ii) parameter-projection normalization, which controls the effective function class of the actor and critic; and (iii) critic multiplicity, embodied by clipped double-Q versus single-Q targets. 
Holding the training backbone, optimizer, environment interface, and network fixed, we vary each axis independently and measure its effect under data-limited (CPU-scale) versus data-abundant (GPU-parallel) regimes across eight benchmark families, 67 environments/tasks in total: DeepMind Control Suite hard tasks~\citep{tassa2018deepmind}, HumanoidBench~\citep{sferrazza2024humanoidbench}, Gym--MuJoCo~\citep{todorov2012mujoco}, MyoSuite~\citep{caggiano2022myosuite}, MuJoCo Playground~\citep{zakka2025mujoco}, MJLab~\citep{zakka2026mjlab}, IsaacLab~\citep{mittal2025isaaclab}, and ManiSkill~\citep{taomaniskill3}. 
The analysis yields a consistent message: \textbf{these stabilizers are data-regime-dependent, not universal}. 
Normalization helps when replay is narrow but restricts value fitting when replay is broad; clipped double-Q can be relaxed in high-throughput manipulation without degrading performance; SWD, in contrast, remains useful across both regimes.

Turning this analysis into a prescription, we build \method, a regime-aware family of off-policy RL algorithm.
\method{} uses SWD as a consistent component and matches each stabilizer choice to the target data regime: \textbf{\method{}-L} (Norm ON, double-Q) for data-limited CPU-scale training, and \textbf{\method{}-A} (Norm OFF, Single-Q) for GPU-parallel training.
Across nine CPU-scale environments, \method{} improves mean normalized score--step AUC over FlashSAC by $4.5\%$; across fourteen GPU-parallel environments, it improves the corresponding AUC by $23.1\%$. On UnitreeG1TransportBox-v1, \method{}-A lifts success rate from $19.8\%$ to $96.4\%$, and on MuJoCo Playground it gains $19.1\%$ in mean normalized wall-time AUC. 
Moreover, we show that \method{} delivers a sim-to-real training-deployment closed-loop for \texttt{Unitree G1} in 35 minutes on a single A800 GPU, showing a 36.4\% wall-clock time reduction compared to FlashSAC. 
This points toward a in-minutes deployment with \method{} when a high-end GPU (e.g., B200) is available.
These gains are obtained without changing the training backbone, without adding auxiliary networks, and often while removing components, showing that a regime-matched selection of stabilizers strictly outperforms uniform inheritance.

Our contributions are threefold:
\begin{itemize}
    \item \textbf{A regime-aware design principle for scalable off-policy RL.} Through controlled component-wise ablations, we show that parameter normalization and clipped double-Q are data-regime-dependent stabilizers, whereas age-biased replay weighting is broadly beneficial. This reframes scalable off-policy RL as a data-regime-matching problem rather than a stabilizer-stacking problem.
    \item \textbf{\method, a regime-aware refinement of FlashSAC that wins across regimes.} \method{} makes better exploitation of data with SWD and pairs each regime with the stabilizer configuration based on our empirical analysis: \method{}-L for CPU-scale training and \method{}-A for GPU-parallel training, all under a shared training backbone.
    \item \textbf{State-of-the-art results across eight benchmark families.} \method{} improves normalized score--step AUC over FlashSAC by $4.5\%$ across nine CPU-scale environments and $23.1\%$ across fourteen GPU-parallel environments, lifts UnitreeG1TransportBox-v1 success rate from $19.8\%$ to $96.4\%$, and gains $19.1\%$ in mean normalized wall-time AUC on MuJoCo Playground.

\end{itemize}

\section{Related Work}
\label{sec:related}

\vspace{-0.3cm}
\paragraph{Scalable off-policy reinforcement learning.}
Off-policy reinforcement learning is attractive for continuous control because it can reuse past experience through replay, improving data efficiency compared with purely on-policy methods. Soft Actor-Critic (SAC) combines off-policy actor--critic learning with entropy regularization and has become a standard baseline for continuous control~\citep{haarnoja2018sac,haarnoja2018applications}. Subsequent methods improve stability or sample efficiency by reducing value-estimation error, increasing update-to-data ratios, using critic ensembles, or learning latent dynamics models~\citep{fujimoto2018td3,chen2021redq,hansen2024tdmpc2}. Recent scalable robot-learning systems further show that off-policy RL can benefit from high-throughput simulation, large replay buffers, and larger neural networks~\citep{lee2025hyperspherical,kim2026flashsac}. Our work follows this scalable off-policy direction, but studies a different question: which classical stabilizers remain useful when replay coverage is broadened by massively parallel data collection.

\vspace{-0.3cm}
\paragraph{Replay weighting, data reuse, and exploitation.}
Replay buffers are commonly sampled uniformly, but non-uniform replay has long been used to improve data reuse. Prioritized experience replay samples transitions according to temporal-difference error, emphasizing transitions with larger learning signals~\citep{schaul2016prioritized}. More recent work studies replay through the lens of non-stationarity and plasticity loss. Deep RL agents can lose adaptability during training due to primacy bias, dormant neurons, rank collapse, weakened gradient signals, or severe value and policy churn~\citep{nikishin2022primacy,sokar2023redo,wu2026rank,tang2024chain,Tang25CCHAIN}. Sample Weight Decay (SWD) addresses this issue by reweighting samples according to age~\citep{wu2026rank}. While SWD was originally motivated by plasticity preservation, we interpret it as an exploitation-oriented replay mechanism: it changes how collected data are reused, rather than adding additional exploration or stronger conservatism.

\vspace{-0.3cm}
\paragraph{Critic conservatism and clipped double-Q learning.}
A central challenge in off-policy actor--critic learning is value-estimation error. TD3 and SAC use clipped double-Q targets to reduce overestimation by taking the minimum over two target critics~\citep{fujimoto2018td3,haarnoja2018sac}. This conservative target is especially useful when the critic must evaluate actions that are poorly covered by the replay buffer, where extrapolation error can otherwise produce spuriously high values. Such kind of extrapolation errors are addressed from different perspectives, e.g., reining the representation-level generalization~\citep{ma2023reining}, reducing out-of-batch churn~\citep{tang2024chain}. However, this conservatism also introduces bias: when data coverage is broad and value fitting becomes the dominant bottleneck, clipped double-Q targets may become overly pessimistic or computationally redundant. Our single-Q experiments examine this trade-off in scalable robotic manipulation settings, asking whether critic-side conservatism can be relaxed when massively parallel simulation already provides diverse replay data.

\vspace{-0.3cm}
\paragraph{Normalization and function-class control.}
Normalization and norm control are widely used to stabilize deep networks. Batch normalization, layer normalization, weight normalization, spectral normalization, and RMS normalization control different aspects of optimization, activation statistics, or parameter geometry~\citep{ioffe2015batch,ba2016layer,salimans2016weight,miyato2018spectral,zhang2019rmsnorm}. In reinforcement learning, norm control is particularly relevant because bootstrapped targets and non-stationary replay can amplify value-function instability. Recent architectures such as SimbaV2 use hyperspherical normalization to improve scalable RL, while XQC studies how critic conditioning affects optimization geometry~\citep{lee2025hyperspherical,palenicek2026xqc}. Our work is complementary: rather than proposing a new normalization layer, we study when parameter projection normalization should be retained or relaxed. We find that norm control is useful in data-limited regimes, but can restrict exploitation in large-parallel regimes where replay coverage is already broad.

\vspace{-0.3cm}
\paragraph{Exploration complexity and data coverage.}
Theoretical analyses of exploration often relate sample complexity to uncertainty over the value-function class. Eluder dimension measures sequential dependence within a function class and appears in optimistic exploration analyses under function approximation~\citep{russo2013eluder,osband2014model}. Norm and Lipschitz constraints provide one way to control the effective complexity of neural function classes, and spectral-norm-based measures have been used to analyze neural-network generalization~\citep{bartlett2017spectrally,miyato2018spectral}. This perspective motivates parameter projection normalization as an exploration- and stability-oriented mechanism: by controlling the Lipschitz constant, it can reduce harmful extrapolation on under-covered regions. Our empirical results suggest that the value of this control depends on the data regime. When parallel simulation provides broad coverage, the same constraint can reduce expressive freedom and weaken exploitation.

\section{Preliminaries}
\label{sec:preli}

\vspace{-0.3cm}
\paragraph{Markov decision process.}
We consider continuous-control reinforcement learning in a discounted Markov decision process (MDP)
$\mathcal{M}=(\mathcal{S},\mathcal{A},P,r,\gamma)$, where $\mathcal{S}$ and $\mathcal{A}$ denote the state and action spaces, $P(s'|s,a)$ is the transition kernel, $r(s,a)$ is the reward function, and $\gamma\in[0,1)$ is the discount factor. At each timestep, the agent observes $s_t$, samples an action $a_t\sim\pi(\cdot|s_t)$, receives reward $r_t$, and transitions to $s_{t+1}$. We use $\rho_\pi(s,a)$ to denote the discounted state-action visitation distribution induced by policy $\pi$. The standard objective is to learn a policy maximizing the expected discounted return,
\[
J(\pi)=\mathbb{E}_{(s_t,a_t)\sim\rho_\pi}
\left[\sum_{t=0}^{\infty}\gamma^t r(s_t,a_t)\right].
\]

\vspace{-0.3cm}
\paragraph{Soft actor-critic.}
Our study builds on Soft Actor-Critic (SAC), an off-policy actor-critic algorithm under the maximum-entropy RL framework~\citep{haarnoja2018sac,haarnoja2018applications}. SAC augments the return objective with an entropy term,
\[
J_{\mathrm{SAC}}(\pi)=
\mathbb{E}_{(s_t,a_t)\sim\rho_\pi}
\left[
\sum_{t=0}^{\infty}\gamma^t
\left(r(s_t,a_t)+\alpha \mathcal{H}(\pi(\cdot|s_t))\right)
\right],
\]
where $\alpha$ controls the relative weight of entropy. In practice, SAC maintains a replay buffer $\mathcal{D}$, a stochastic policy $\pi_\theta$, and two critics $Q_{\phi_1},Q_{\phi_2}$ to counter overestimation bias via the \emph{clipped double-Q target}~\citep{fujimoto2018td3}: given transitions $(s,a,r,s')\sim\mathcal{D}$ and $a'\sim\pi_\theta(\cdot|s')$,
\[
y=r+\gamma\left(
\min_{i=1,2}Q_{\bar{\phi}_i}(s',a')
-\alpha \log \pi_\theta(a'|s')
\right),
\]
and each critic is trained by minimizing the Bellman error
\[
\mathcal{L}_{Q_i}(\phi_i)
=
\mathbb{E}_{(s,a,r,s')\sim\mathcal{D}}
\left[
\left(Q_{\phi_i}(s,a)-y\right)^2
\right].
\]
The clipped double-Q operator suppresses spuriously high $Q$ estimates on poorly covered actions but adds a second critic and a pessimism bias; a \emph{single-Q} variant simply removes the minimization and uses one critic.

FlashSAC extends this SAC foundation to large-scale robotic control by combining high-throughput data collection, larger models, reduced update frequency, and norm-control mechanisms for stable critic learning~\citep{kim2026flashsac}.
Concretely, high-throughput parallel simulation together with a low gradient-update-to-environment-step ratio shifts wall-clock cost from gradient computation to environment interaction, making larger actor--critic networks affordable at scale; parameter projection normalization then keeps such large critics stable by column-wise renormalizing linear weights after each optimizer step, which constrains the effective function class and controls value extrapolation on poorly covered actions.

Together, these design choices make FlashSAC a strong scalable off-policy backbone on continuous control, on which we build \method{} in the remainder of the paper.

\vspace{-0.3cm}
\paragraph{Parameter projection normalization.}
FlashSAC includes norm-control components to stabilize large-scale off-policy learning~\citep{kim2026flashsac}. We focus on parameter projection normalization as a controllable factor. Consider a neural network $f_\theta$ with weight matrices $\{W_\ell\}_{\ell=1}^{L}$. Parameter projection constrains each layer after optimization, for example by projecting $W_\ell$ into a Frobenius-norm ball,
\[
W_\ell \leftarrow \Pi_{\|W\|_F\le c_\ell}(W_\ell).
\]
Since $\|W_\ell\|_2\le \|W_\ell\|_F$, bounding the Frobenius norm also bounds the spectral norm. For networks with 1-Lipschitz activations, this yields the standard Lipschitz upper bound
\[
\mathrm{Lip}(f_\theta)
\le
\prod_{\ell=1}^{L}\|W_\ell\|_2
\le
\prod_{\ell=1}^{L}c_\ell .
\]
This connects parameter projection normalization to spectral-norm-based complexity control in neural networks~\citep{bartlett2017spectrally,miyato2018spectral}. In our analysis, this normalization is treated as an exploration- and stability-oriented constraint: it controls the effective function class, but may also restrict expressive freedom when the replay distribution already provides sufficient data coverage.

\vspace{-0.3cm}
\paragraph{Weighted replay.}
Uniform replay samples each stored transition with equal probability. More generally, let $\mathcal{D}_t=\{\tau_i\}_{i=1}^{N_t}$ be the replay buffer at training step $t$, where $\tau_i=(s_i,a_i,r_i,s'_i)$ denotes a transition. A weighted-replay scheme samples $\tau_i$ with probability
\[
p_t(i)=\frac{w_t(i)}{\sum_{j=1}^{N_t}w_t(j)},
\]
where $w_t(i)\ge 0$ is a sample weight. Under weighted replay, the critic objective becomes
\[
\mathcal{L}_{Q_i}^{w}(\phi_i)
=
\mathbb{E}_{\tau_i\sim p_t}
\left[
\left(Q_{\phi_i}(s_i,a_i)-y_i\right)^2
\right].
\]
Different choices of $w_t(i)$ recover prior schemes such as uniform sampling and prioritized replay~\citep{schaul2016prioritized}. 
Sample Weight Decay (SWD) instantiates this idea by assigning age-aware weights to replay samples, and was proposed as a lightweight replay-side method for mitigating plasticity loss in deep RL~\citep{wu2026rank}. In this paper, we study SWD not only as a plasticity-preserving method, but also as a mechanism that facilitates data exploitation during off-policy learning.

\vspace{-0.2cm}
\paragraph{Eluder dimension and exploration complexity.}
Eluder dimension measures the degree of sequential dependence within a function class and is commonly used to characterize exploration complexity under function approximation~\citep{russo2013eluder,osband2014model}. Intuitively, a function class with lower eluder dimension requires fewer informative observations to resolve uncertainty about future predictions. Norm and Lipschitz constraints can reduce the effective complexity of the value-function class, providing a theoretical lens for understanding why parameter projection normalization may support exploration and stable generalization. We use this connection as a conceptual interpretation rather than a formal sample-complexity guarantee for the neural networks studied empirically.

\section{WarpSAC}
\label{sec:rainbowsac}

The preceding sections show that classical off-policy stabilizers carry implicit data-regime assumptions: their benefits depend on whether replay coverage is narrow or broad.
Rather than applying them uniformly, we propose \method{}, a regime-aware family built on FlashSAC that explicitly matches each stabilizer choice to the available data coverage.
Concretely, \method{} varies three axes: replay weighting~$w_t(i)$, parameter projection normalization, and critic multiplicity, and pairs different settings of these axes with different data regimes.
The rest of this section states the underlying hypothesis (Section~\ref{sec:regime_hypothesis}), identifies the three axes (Section~\ref{sec:three_axes}), defines the regime-agnostic component SWD (Section~\ref{sec:swd}), and combines them into the concrete \method{} prescription (Section~\ref{sec:prescription}).

\subsection{Data-Regime Hypothesis}
\label{sec:regime_hypothesis}

Off-policy RL stabilizers such as clipped double-Q, parameter projection normalization, and uniform replay were developed under CPU-scale assumptions, where replay coverage is narrow and value extrapolation is fragile.
GPU-parallel simulation changes the data regime: thousands of actors populate the buffer with diverse trajectories at high throughput, so the bottleneck shifts from obtaining sufficient coverage to fitting and exploiting high-value behavior from abundant data.
We hypothesize that this shift changes the relative value of the classical stabilizers.
In the \emph{data-limited} regime, parameter normalization and clipped double-Q help by constraining the effective function class and suppressing spuriously high $Q$ values on poorly covered actions, while any replay-side mechanism should make better use of the narrow buffer.
In the \emph{data-abundant} regime, the same conservative mechanisms can restrict value fitting or add unnecessary pessimism, while replay-side gains persist because targeting policy-relevant transitions is orthogonal to coverage.

\subsection{The Three Axes}
\label{sec:three_axes}

Given this hypothesis, which design knobs most directly encode regime assumptions? We isolate three, all on top of the FlashSAC~\citep{kim2026flashsac}.
\method{} varies only these axes:
\begin{itemize}[leftmargin=3em,topsep=2pt,itemsep=2pt]
    \item[\axisA] \textbf{Replay weighting $w_t(i)$.} Whether transitions in the buffer are sampled uniformly or with an age-dependent weight (Section~\ref{sec:swd}).
    \item[\axisB] \textbf{Parameter projection normalization.} Whether the FlashSAC column-wise weight renormalization is applied after each optimizer step (\emph{Norm ON}) or disabled (\emph{Norm OFF}).
    \item[\axisC] \textbf{Critic multiplicity.} Whether the clipped double-Q target is used (two critics, the FlashSAC default) or replaced by a single critic (\emph{Single-Q}).
\end{itemize}
We fix data-collection throughput and optimizer schedule to compare regimes cleanly, and vary network capacity only in the separate scale ablation (Section~\ref{sec:experiments}).
Varying \axisA--\axisC\ independently is what enables regime-aware pairing rather than collapsing \method{} into a single ``strong'' or ``weak'' SAC recipe.

\subsection{Sample Weight Decay}
\label{sec:swd}

Among the three axes, replay weighting is the only one we apply regardless of data regime. SWD instantiates \axisA: it biases minibatch sampling toward recent transitions using a linear age decay. Concretely, let $\tau_i=(s_i,a_i,r_i,s'_i)$ be a transition inserted into the replay buffer at time $t_i$, and let $A_t(i)=t-t_i$ be its age at training step~$t$. SWD assigns each transition the age-dependent weight
\begin{equation}
    w_t(i) = \max \!\left( w_{\min},\ 1 - \tfrac{A_t(i)}{T_{\mathrm{decay}}} \right),
    \qquad
    p_t(i) = \tfrac{w_t(i)}{\sum_j w_t(j)},
\end{equation}
where $T_{\mathrm{decay}}$ is the decay horizon and $w_{\min}>0$ is a floor that prevents old transitions from being fully discarded. Setting $T_{\mathrm{decay}}=0$ recovers uniform replay. SWD is implemented inside the replay buffer with no auxiliary networks, additional Bellman targets, or loss changes.

Policy performance is dominated by Bellman errors on state--action regions visited by the current policy and on regions along high-value trajectories~\citep{wu2026rank}. Uniform replay ignores this structure and spends equal update probability on transitions from much older policies. SWD redirects a fixed update budget toward more policy-relevant transitions by changing only the minibatch distribution, without touching the nominal UTD or the update rule; a nonzero floor $w_{\min}$ preserves coverage. Because this argument depends on \emph{policy age}, not on data volume, we retain SWD in both regimes, which serves as a consistent design choice confirmed empirically in Section~\ref{sec:experiments}.

Policy performance is dominated by Bellman errors on state--action regions visited by the current policy and on regions along high-value trajectories~\citep{wu2026rank}. Uniform replay ignores this structure and spends equal update probability on transitions from much older policies. SWD redirects a fixed update budget toward more policy-relevant transitions by changing only the minibatch distribution, without touching the nominal UTD or the update rule; a nonzero floor $w_{\min}$ preserves coverage. Because this argument depends on \emph{policy age}, not on data volume, we retain SWD in both regimes---a design choice confirmed empirically in Section~\ref{sec:experiments}.

\subsection{Regime-Aware Variants and Prescription}
\label{sec:prescription}

With SWD established as the regime-agnostic replay component, the remaining design question is how to set the other two axes, i.e., normalization and critic multiplicity, for each data regime. Table~\ref{tab:prescription} summarizes the resulting variants. \method{}-L and \method{}-A are the two prescribed variants for their respective regimes; \method{} w Norm OFF serves as an intermediate ablation point. FlashSAC (no SWD, Norm ON, clipped double-Q) is the shared baseline.

These variants form a conservatism spectrum---from full (Norm ON, double-Q) to minimal (Norm OFF, single-Q)---with SWD applied throughout, enabling controlled comparison across data regimes.
A key feature of \method{} is that the GPU-parallel recipe achieves gains by \emph{removing} components rather than adding them: dropping normalization frees the critic to fit abundant data, and dropping the second critic halves critic-side computation. The result is an algorithm that is both simpler and stronger than the fully stabilized baseline. This stands in contrast to the common pattern of stacking mechanisms for robustness; \method{} shows that matching stabilizers to the data regime is more effective than uniform inheritance.

\begin{table}[ht]
\centering
\caption{\textbf{\method{} regime-aware prescription.} SWD is applied in all variants. Normalization and critic multiplicity are set according to the data regime. FlashSAC (no SWD, Norm ON, double-Q) serves as the shared baseline.}
\label{tab:prescription}
\scalebox{0.9}{%
\setlength{\tabcolsep}{5pt}
\renewcommand{\arraystretch}{1.15}
\begin{tabular}{l|l|cccc}
\toprule
Data Regime & Variant Name & Replay \axisA & Normalization \axisB & Critic \axisC \\
\midrule
Data-limited (CPU-scale) & \method{}-L & SWD & Norm ON & Clipped Double-Q \\
Data-abundant (GPU-parallel) & \method{}-A & SWD & Norm OFF & Single-Q \\
\midrule
Ablation & \method{} w Norm OFF & SWD & Norm OFF & Clipped Double-Q \\
\bottomrule
\end{tabular}%
}
\end{table}

In short, \method{} is not a single algorithm but a design principle: the same three-axis framework produces different concrete recipes for different data regimes, and the experiments in Section~\ref{sec:experiments} confirm that this regime-aware selection consistently outperforms uniform stabilizer inheritance.

\begin{tcolorbox}[
    colback=axisbluebg!30,
    colframe=accentcolor,
    boxrule=0.8pt,
    arc=2pt,
    left=1.0em,
    right=1.0em,
    top=0.8em,
    bottom=0.8em,
    title={\rule[-3pt]{0pt}{12pt}\textbf{Practitioner's Guide: Choosing a \method{} Variant}},
    fonttitle=\sffamily\small,
    coltitle=white,
    colbacktitle=accentcolor
]
\small
\begin{itemize}[leftmargin=1.2em,itemsep=4pt,topsep=2pt]
    \item \textbf{Use} \method{}-L\, \textbf{for Data-limited (CPU-scale) Regime}: Normalization and conservatism stabilizes value extrapolation under narrow replay coverage.
    \item \textbf{Use} \method{}-A\, \textbf{for Data-abundant (GPU-parallel) Regime}: Removing normalization and conservatism frees the critic to exploit broad, rapidly refreshed replay data.
    \item \textbf{Use SWD in Both Regimes}: Always enable SWD. Age-aware replay improves update efficiency regardless of data volume.
\end{itemize}
\end{tcolorbox}

\section{Experiments}
\label{sec:experiments}

Earlier sections introduced the data-regime hypothesis and the \method{} family, which studies how canonical off-policy RL configurations should adapt to a shift in data regime: components that help under limited replay may become restrictive when data are abundant.
In the experiments, we evaluate this hypothesis across eight benchmark families spanning CPU-scale and massively parallel GPU training, organizing the experiments around five questions.
Sections~\ref{sec:cpu} and~\ref{sec:gpu} address Q1--Q3 in the two regimes.
Further, we move on from simulation to real-world evaluation of \method{} for \texttt{Unitree-G1} to address Q4 in Section~\ref{sec:sim2real}.
Finally, we delve into the mechanism analysis to address Q5 in
Sections~\ref{sec:ablation_capacity} and~\ref{sec:ablation_norm}.

The research questions to answer by our experiments are listed below.
\begin{tcolorbox}[
    colback=preprintfg!6!white,
    colframe=preprintfg!20,
    boxrule=0.5pt,
    arc=2pt,
    left=0.8em, right=0.8em, top=0.6em, bottom=0.6em
]
\begin{itemize}[leftmargin=2.4em,itemsep=3pt,topsep=2pt]
    \item[\textbf{Q1.}] Do the \method{} variants excel in their intended regimes, with \method{}-L targeting data-limited settings and \method{}-A targeting data-abundant settings?
    \item[\textbf{Q2.}] Does SWD remain beneficial as the regime-agnostic component across both regimes?
    \item[\textbf{Q3.}] Do conservative stabilizers, namely normalization and clipped double-Q, shift from beneficial to restrictive as data scale increases?
    \item[\textbf{Q4.}] Can \method{} achieve faster sim-to-real training than FlashSAC under the same setup?
    \item[\textbf{Q5.}] How does replay weighting interact with network capacity across the two regimes?
\end{itemize}
\end{tcolorbox}

\subsection{Experimental Setup}

We evaluate \method{} across eight benchmark families under two operational data-collection regimes:
\begin{itemize}[leftmargin=1.5em,itemsep=2pt,topsep=3pt]
    \item \textbf{Data-limited (CPU-scale):} MuJoCo \citep{todorov2012mujoco}, DeepMind Control Suite hard tasks \citep{tassa2018deepmind}, HumanoidBench \citep{sferrazza2024humanoidbench}, and MyoSuite \citep{caggiano2022myosuite}.
    \item \textbf{Data-abundant (GPU-parallel):} MuJoCo Playground \citep{zakka2025mujoco}, IsaacLab \citep{mittal2025isaaclab}, MJLab \citep{zakka2026mjlab}, and ManiSkill \citep{taomaniskill3}.
\end{itemize}
These domains cover standard locomotion, humanoid control, dexterous manipulation, and massively parallel GPU simulation.
For return-based domains, we report episodic return; for manipulation domains with sparse success signals (ManiSkill), we report average success metrics following the benchmark convention.

We compare four variants.
\textbf{FlashSAC} is the uniform-replay baseline (no SWD, Norm ON, clipped double-Q).
\textbf{\method{}-L} adds SWD while keeping normalization and clipped double-Q enabled.
\textbf{\method{} w Norm OFF} adds SWD and disables parameter projection normalization, retaining clipped double-Q.
For GPU-parallel domains, we additionally evaluate \textbf{\method{}-A}, which further removes clipped double-Q and uses a single critic.
All variants share the same training backbone, optimizer, environment interface, and logging pipeline; only the three axes \axisA--\axisC\ are varied.
Full hyperparameters, network architecture, and training details are provided in Appendix~\ref{app:experiments}.



\subsection{Data-Limited Regime: CPU-Scale Environments}
\label{sec:cpu}

\begin{figure}[H]
    \vspace{-0.3cm}
    \centering
    \includegraphics[width=1\linewidth]{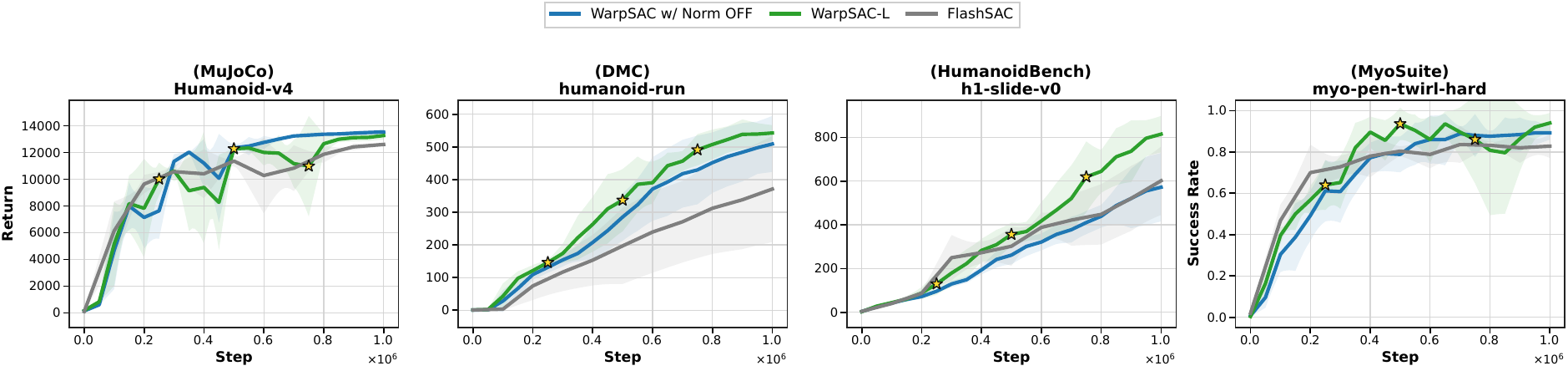}
    \vspace{-0.1cm}
    \caption{\textbf{Learning curves of \method{} and FlashSAC in the data-limited regime (CPU-scale).}
    Return (mean $\pm$ std over five seeds) versus environment steps for FlashSAC, \method{}-L, and \method{} w/ Norm OFF on representative tasks from MuJoCo, DMC, HumanoidBench, and MyoSuite.
    Both \method{} variants clearly surpass FlashSAC on all four tasks, and \method{}-L attains the strongest final return on humanoid-run, h1-slide-v0, and myo-pen-twirl-hard.
    }
    \label{fig:cpu_sim}
    \vspace{-0.3cm}
\end{figure}

We first examine representative CPU-scale environments where replay coverage is relatively limited and value extrapolation is more likely to affect learning stability.
Figure~\ref{fig:cpu_sim} compares FlashSAC with \method{} variants on MuJoCo, DMC hard tasks, HumanoidBench, and MyoSuite tasks.
These environments cover standard locomotion, humanoid control, whole-body control, and dexterous manipulation, while avoiding massively parallel GPU data collection.

Across these tasks, SWD consistently improves over FlashSAC or reaches stronger final performance, suggesting that age-aware replay weighting improves the effective use of collected data.
On humanoid-run, h1-slide-v0, and myo-pen-twirl-hard, the normalized \method{} variant achieves the strongest final performance and often learns more stably.
This supports our hypothesis that parameter projection normalization is useful when replay coverage is narrower: by constraining the effective function class, normalization helps control value extrapolation and stabilizes actor--critic updates.
On Humanoid-v4, the norm-off variant reaches a slightly higher final return, but both \method{} variants remain clearly above FlashSAC, indicating that the replay-side benefit of SWD is robust even when the best normalization choice is task-dependent.

Overall, the CPU-scale results show that \method{}-L, i.e., the prescribed data-limited variant, is the strongest configuration (Q1), driven by SWD's broad benefit across all tasks (Q2) and normalization's role in constraining value extrapolation under limited replay (Q3).
Rather than treating normalization as universally beneficial, these results suggest that its role is tied to the amount and diversity of replay data available to the learner.

\begin{tcolorbox}[
    colback=yellow!10,
    colframe=yellow!50!orange,
    boxrule=0.5pt,
    arc=2pt,
    left=0.8em, right=0.8em, top=0.5em, bottom=0.5em,
    title={\rule[-3pt]{0pt}{12pt}\textbf{Takeaway 1 --- In the data-limited regime, \method{}-L excels by pairing SWD with normalization}},
    fonttitle=\sffamily\small,
    coltitle=black,
    colbacktitle=yellow!25
]
\small
\method{}-L, i.e., the prescribed data-limited variant, achieves the strongest or most stable performance, confirming the effectiveness of pairing SWD with normalization suits the limited-replay regime.
\end{tcolorbox}

\subsection{Data-Abundant Regime: GPU-Parallel Environments}
\label{sec:gpu}

\begin{figure}[H]
    \vspace{-0.1cm}
    \centering
    \includegraphics[width=1\linewidth]{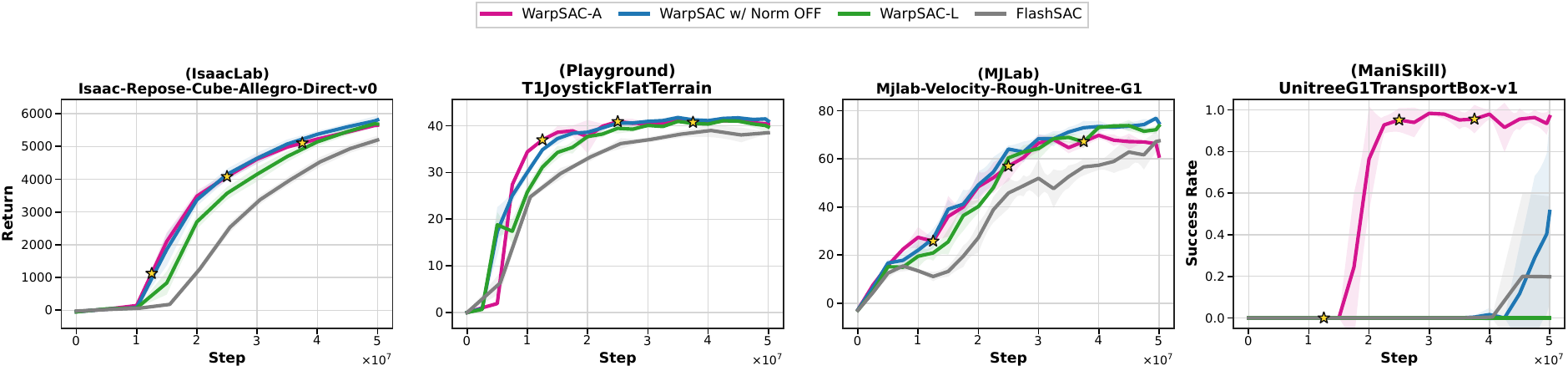}
    \caption{\textbf{Learning curves of \method{} and FlashSAC in the data-abundant regime (GPU-parallel).}
    Return (mean $\pm$ std over five seeds) versus environment steps for FlashSAC, \method{}-L, \method w/ Norm OFF, and \method{}-A on representative tasks from IsaacLab, MuJoCo Playground, MJLab, and ManiSkill.
    \method{} w/ Norm OFF matches or outperforms \method{}-L on IsaacLab, MuJoCo Playground, and MJLab, and \method{}-A attains the strongest final return on ManiSkill.
    }
    \label{fig:gpu_sim}
    \vspace{-0.3cm}
\end{figure}

We turn to GPU-parallel environments, where many actors collect data simultaneously and the replay buffer is refreshed with broad state--action coverage.
Figure~\ref{fig:gpu_sim} compares FlashSAC, \method{} with and without parameter normalization, and a single-Q norm-off variant on IsaacLab, MuJoCo Playground, MJLab, and ManiSkill.
This comparison tests whether conservative SAC components remain necessary when scalable simulation already provides diverse replay data.

The results show a different pattern from the CPU-scale setting.
In IsaacLab, MuJoCo Playground, and MJLab, the norm-off variant is competitive with or stronger than the normalized variant, indicating that parameter projection normalization can become restrictive when the replay buffer already provides sufficient coverage.
On the MJLab rough-terrain Unitree G1 task, the two-critic \method{} variants outperform the uniform-replay FlashSAC baseline, while \method{}-A is less consistent.
This shows that critic-side conservatism remains task-dependent even within GPU-parallel training.
In ManiSkill, the single-Q norm-off variant achieves the strongest performance, suggesting that clipped double-Q conservatism can be relaxed in some high-throughput manipulation settings without degrading learning.
This supports the interpretation that, in GPU-parallel regimes, the bottleneck shifts from conservative generalization under scarce data to efficient exploitation of abundant data.

These results confirm that \method{}-A, i.e., the prescribed data-abundant variant, is a viable and often stronger choice at scale (Q1), because normalization and clipped double-Q shift from beneficial to restrictive as data volume increases (Q3).
When parallel simulation supplies broad replay coverage, strong conservative biases may reduce the flexibility of the value function or add unnecessary computation.
SWD remains useful across both regimes because it reallocates updates toward policy-relevant samples without changing the actor--critic objective.

\begin{tcolorbox}[
    colback=yellow!10,
    colframe=yellow!50!orange,
    boxrule=0.5pt,
    arc=2pt,
    left=0.8em, right=0.8em, top=0.5em, bottom=0.5em,
    title={\rule[-3pt]{0pt}{12pt}\textbf{Takeaway 2 --- In the data-abundant regime, \method{}-A excels by lifting conservatism burdens}},
    fonttitle=\sffamily\small,
    coltitle=black,
    colbacktitle=yellow!25
]
\small
Conservative stabilizers (i.e., normalization, clipped double-Q) shift from beneficial to restrictive as data scale increases. \method{}-A validates that relaxing both is viable in data-abundant settings, while SWD remains useful across regimes.
\end{tcolorbox}

\subsection{Sim-to-Real Evaluation: Training Unitree G1 to Walk in Real World}\label{sec:sim2real}

Off-policy RL is often viewed as brittle for sim-to-real transfer, especially on high-dimensional humanoid systems where unstable value learning can quickly produce unsafe locomotion. We therefore include a sim-to-real case study on \texttt{Unitree-G1-Flat}, a 29-DoF Unitree G1 locomotion task with proprioceptive policy inputs from Unitree Robotics' \href{https://github.com/unitreerobotics/unitree_rl_mjlab}{\texttt{unitree\_rl\_mjlab}} pipeline. The task uses an asymmetric actor--critic formulation: the deployed policy receives a 98-dimensional actor observation, while the critic is trained with a 211-dimensional privileged observation. We keep the environment, reward design, observation interface, and sim-to-real adaptation pipeline fixed across methods, and compare \method{}, FlashSAC, and PPO under the same Unitree-G1-Flat training profile.

For PPO, we start from the \texttt{rsl\_rl}-style on-policy pipeline commonly used in Unitree sim-to-real locomotion. The unmodified \texttt{rsl\_rl} implementation does not include the mixed-precision and compilation path used by our scalable off-policy learners. To make the wall-clock comparison less dependent on an avoidably slow baseline implementation, we add bfloat16 autocasting and \texttt{torch.compile} to the PPO training path. Thus, PPO is a strengthened baseline rather than the vanilla \texttt{rsl\_rl} code path.

\begin{figure}[H]
    \centering
    \vspace{-0.2cm}
    \includegraphics[width=0.8\linewidth]{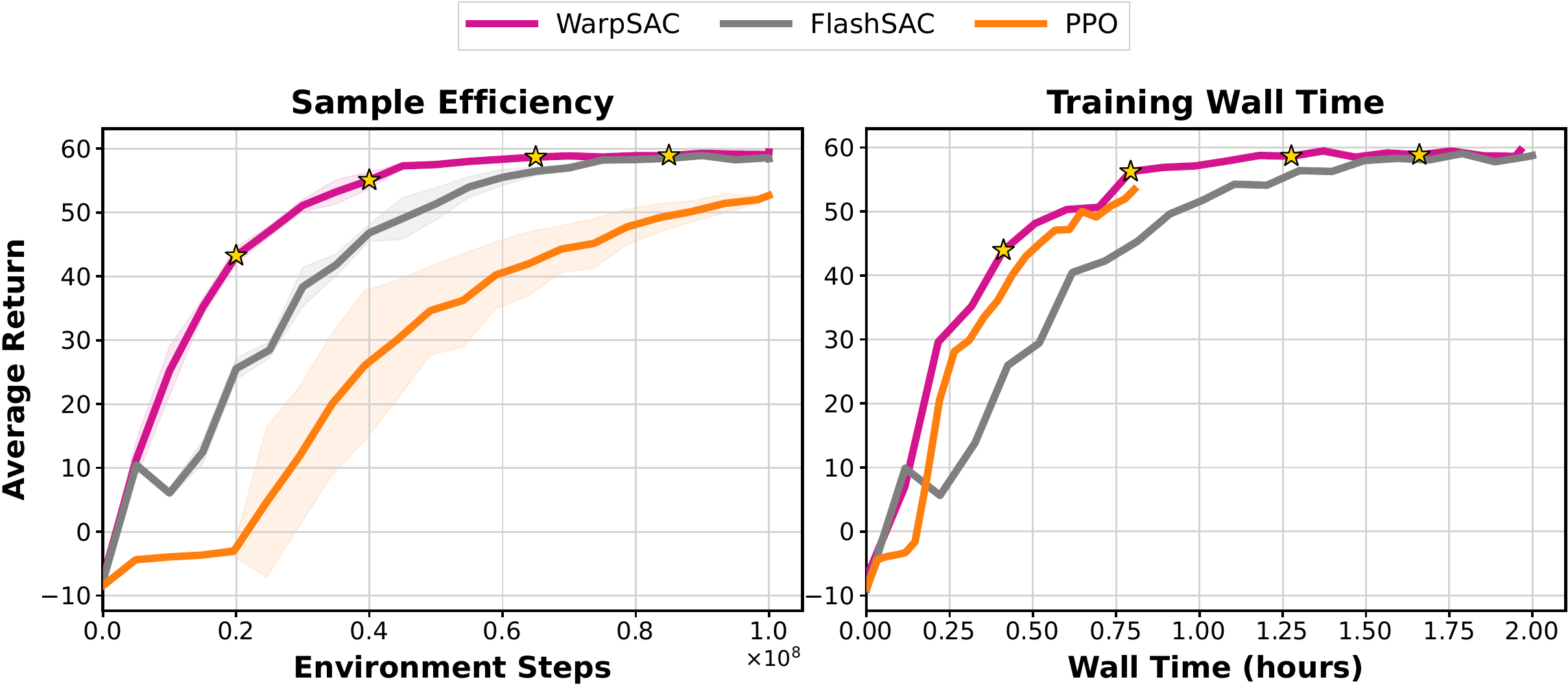}
    \vspace{-0.3cm}
    \caption{\textbf{Unitree G1 Sim-to-Real learning curves.}
    The left panel compares sample efficiency with standard-deviation shading. The right panel compares training wall time.}
    \label{fig:sim2real_unitree_g1_flat}
    \vspace{-0.4cm}
\end{figure}

Figure~\ref{fig:sim2real_unitree_g1_flat} reports both sample-efficiency and wall-time learning curves, and Figure~\ref{fig:intro} also shows the screenshots of the gaits. The video of our real-world deployment can be found on the \href{https://wzhhasadream.github.io/WarpSAC/}{\texttt{project page}}. On an end-to-end A800 run that includes simulation, replay, learner updates, logging, and evaluation, \method{} reaches the target performance in roughly \textbf{35 minutes}, whereas FlashSAC takes about \textbf{55 minutes} under the same sim-to-real setup. These results suggest that the replay-side exploitation benefits of \method{} transfer beyond simulator-only benchmarks, making scalable off-policy learning practical for high-dimensional humanoid sim-to-real locomotion when paired with a stable adaptation pipeline.

\begin{tcolorbox}[
    colback=yellow!10,
    colframe=yellow!50!orange,
    boxrule=0.5pt,
    arc=2pt,
    left=0.8em, right=0.8em, top=0.5em, bottom=0.5em,
    title={\rule[-3pt]{0pt}{12pt}\textbf{Takeaway 3 --- \method{} enables faster sim-to-real deployment than FlashSAC}},
    fonttitle=\sffamily\small,
    coltitle=black,
    colbacktitle=yellow!25
]
\small
\method{} delivers a sim-to-real training-deployment closed-loop for \texttt{Unitree G1} in 35 minutes on a single A800 GPU, showing a 36.4\% wall-clock time reduction compared to FlashSAC. 
This points toward a in-minutes deployment when a high-end GPU (e.g., B200) is available.
\end{tcolorbox}

\subsection{Mechanism Analysis: SWD × Network Capacity in CPU-Scale}
\label{sec:ablation_capacity}

\begin{figure}[t]
    \vspace{-0.1cm}
    \centering
    \includegraphics[width=1\linewidth]{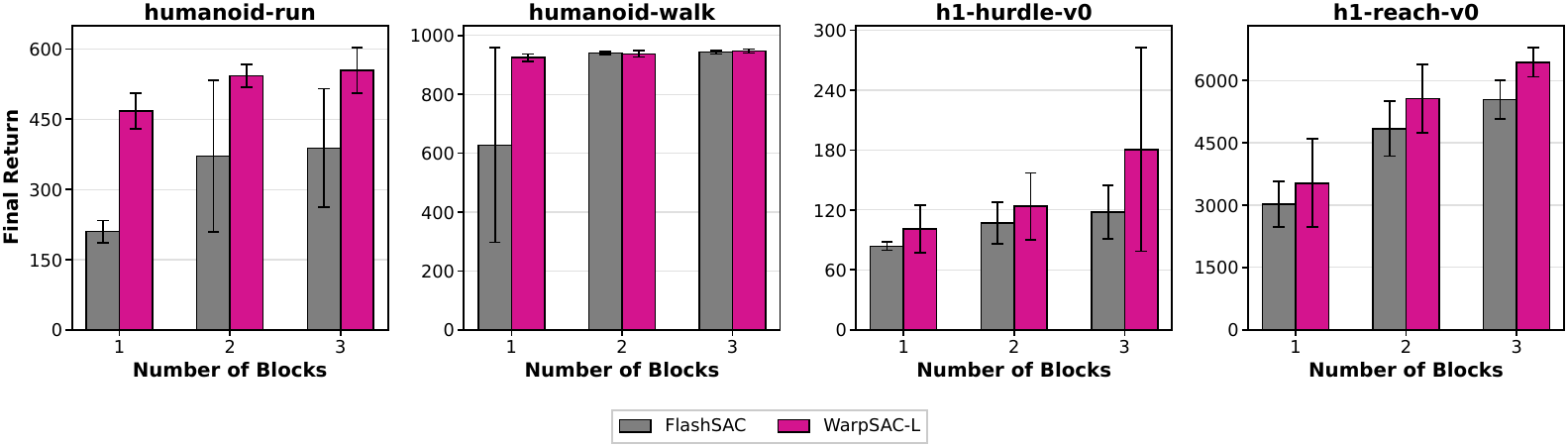}
    \vspace{-0.7cm}
    \caption{\textbf{Network-capacity comparison of \method{}-L and FlashSAC in CPU-scale training.}
    Final return of \method{}-L and FlashSAC as the number of residual blocks is varied on humanoid-run, humanoid-walk, h1-hurdle-v0, and h1-reach-v0. Both methods use Norm ON and clipped double-Q and differ only in replay weighting, isolating how the \method{} replay mechanism interacts with network capacity.
    With a single block, \method{}-L improves final return over FlashSAC across all the four tasks; as blocks increase, the gap narrows on saturated tasks but remains visible on harder HumanoidBench tasks.
    }
    \label{fig:scale_ablation}
    \vspace{-0.2cm}
\end{figure}

To isolate how replay weighting interacts with model capacity (Q4), Figure~\ref{fig:scale_ablation} varies the number of FlashSAC blocks on four CPU-scale locomotion tasks: humanoid-run, humanoid-walk, h1-hurdle-v0, and h1-reach-v0.
SWD is most beneficial in the low-capacity regime: with a single block the relative gain over uniform replay exceeds $2\times$ on humanoid-run (from 209.93 to 467.39) and reaches $\sim$47\% on humanoid-walk (from 627.07 to 924.39), ${\sim}21\%$ on h1-hurdle-v0 (from 83.80 to 101.09), and ${\sim}17\%$ on h1-reach-v0 (from 3018.97 to 3529.44).
These gains suggest that age-biased replay can improve update efficiency by emphasizing more policy-relevant samples when the function approximator is capacity constrained.

As capacity increases, the gap narrows on saturated tasks such as humanoid-walk, while remaining visible on harder HumanoidBench tasks (h1-hurdle-v0, h1-reach-v0).
This trend shows that SWD is not merely a substitute for additional parameters; it changes \emph{how} available capacity is used.
When the network is small, prioritizing recent samples provides a clear optimization advantage; when the network is large enough to fit broader replay distributions, the marginal benefit becomes task-dependent, though SWD remains competitive or better across all tested capacities.

\begin{tcolorbox}[
    colback=yellow!10,
    colframe=yellow!50!orange,
    boxrule=0.5pt,
    arc=2pt,
    left=0.8em, right=0.8em, top=0.5em, bottom=0.5em,
    title={\rule[-3pt]{0pt}{12pt}\textbf{Takeaway 4 --- In CPU-scale training, SWD boosts learning under limited capacity and remains beneficial as networks grow}},
    fonttitle=\sffamily\small,
    coltitle=black,
    colbacktitle=yellow!25
]
\small
SWD compensates for limited network capacity by focusing updates on policy-relevant data. The benefit is largest for small models and narrows (but does not vanish) as capacity grows.
\end{tcolorbox}



\subsection{Mechanism Analysis: Normalization × Network Capacity in GPU-Parallel}
\label{sec:ablation_norm}

\begin{figure}[h]
    \vspace{-0.1cm}
    \centering
    \includegraphics[width=1\linewidth]{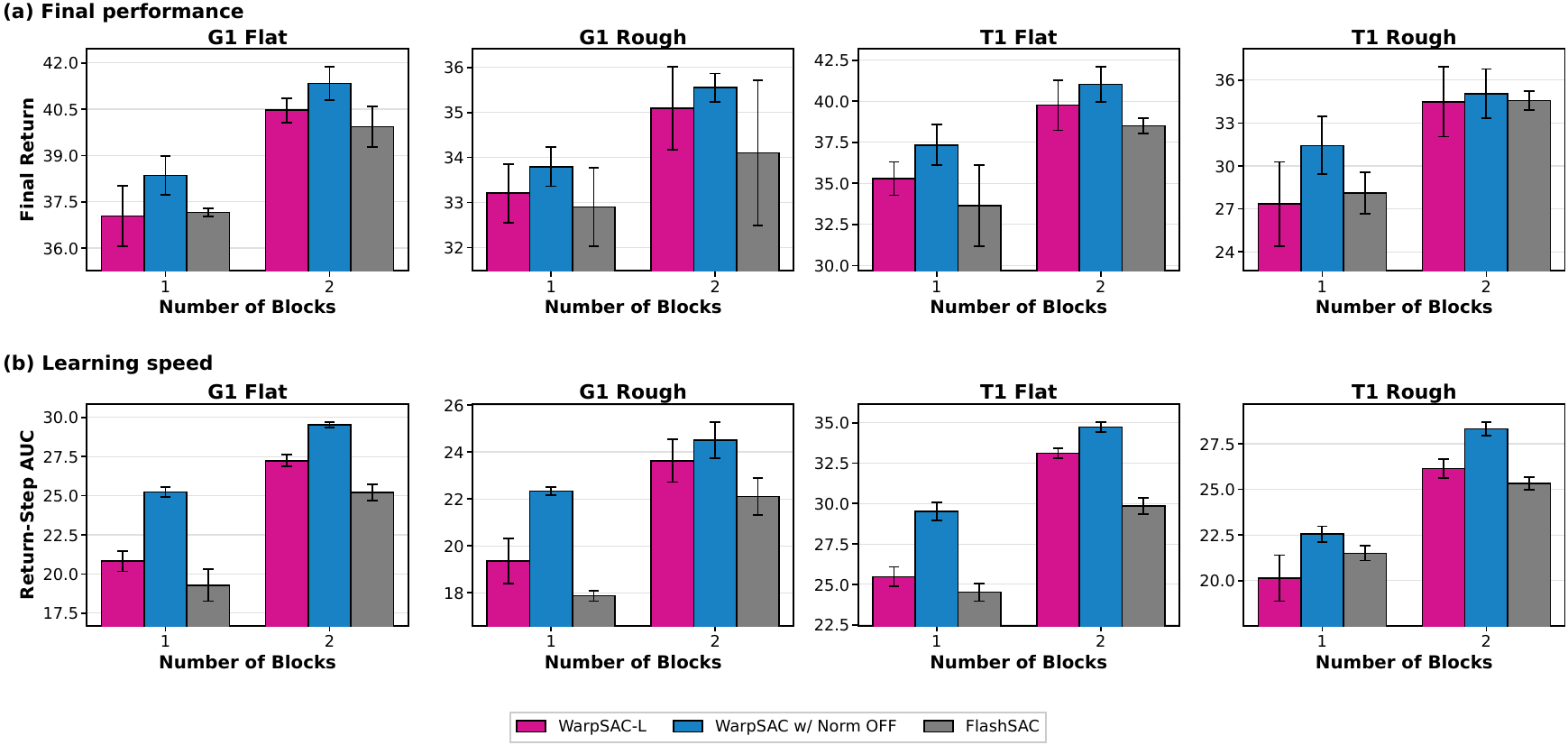}
    \caption{\textbf{Network-capacity comparison of \method{} variants and FlashSAC in GPU-parallel training.}
    Final return and return--step AUC (normalized by the common training horizon, reflecting average return throughout training) of \method{}-L, \method{} w/ Norm OFF, and FlashSAC as the number of residual blocks is varied on MuJoCo Playground tasks.
    At one block, \method{} w/ Norm OFF improves both final return and AUC markedly over \method{}-L and FlashSAC on G1 Flat and T1 Rough; with two blocks the gap narrows, but \method{} w/ Norm OFF remains competitive or better across all four tasks.
    }
    \label{fig:playground_scale}
    \vspace{-0.2cm}
\end{figure}

We study the interaction between SWD, parameter projection normalization, and model capacity in MuJoCo Playground (Figure~\ref{fig:playground_scale}), a massively parallel setting where replay data are generated rapidly and the bottleneck is fitting high-value behavior rather than obtaining sufficient exploration.

The results show that parameter projection normalization can sharply reduce performance when capacity is limited. With one FlashSAC block, disabling normalization improves performance markedly on G1 Flat and T1 Rough. With two blocks, the gap becomes smaller, but Norm OFF remains competitive or better across the Playground tasks. This supports our hypothesis that normalization is useful when data are scarce, but can restrict expressive freedom in large-parallel environments where state--action coverage is already broad.

Importantly, SWD alone is not always sufficient when normalization strongly constrains the function class: the best-performing configurations combine SWD with reduced normalization, confirming that replay-side exploitation and network expressivity must be balanced jointly in GPU-scale RL (Q4).

\begin{tcolorbox}[
    colback=yellow!10,
    colframe=yellow!50!orange,
    boxrule=0.5pt,
    arc=2pt,
    left=0.8em, right=0.8em, top=0.5em, bottom=0.5em,
    title={\rule[-3pt]{0pt}{12pt}\textbf{Takeaway 5 --- In GPU-parallel training, normalization restricts learning at low capacity; pairing SWD with reduced normalization yields the best results}},
    fonttitle=\sffamily\small,
    coltitle=black,
    colbacktitle=yellow!25
]
\small
Disabling normalization improves performance sharply at low capacity (e.g., on G1 Flat and T1 Rough) and remains competitive as capacity grows, indicating that the constraint on the function class outweighs its stabilization benefit when replay coverage is already broad. Pairing SWD with reduced normalization yields the strongest configurations.
\end{tcolorbox}

\section{Conclusion}
\label{sec:conclusion}

Scaling data collection changes the fundamental assumption underlying canonical off-policy RL.
We focus on the data-regime shift and formalized this observation as a \emph{data-regime hypothesis}: the utility of an off-policy stabilizer is not a fixed property of the algorithm, but a function of how well replay coverage matches the class it is designed to protect.
Building on this hypothesis, we proposed \method{}, a regime-aware family of off-policy RL algorithms that treats SWD as a regime-agnostic component and matches conservative stabilizers to the target data regime.
\method{}-L (SWD, Norm ON, double-Q) targets data-limited CPU-scale training, while \method{}-A (SWD, Norm OFF, single-Q) targets data-abundant GPU-parallel training.
Across eight benchmark families spanning both regimes, this prescription is validated by controlled ablations and consistent aggregate gains over FlashSAC.
Among the three studied axes, SWD is the only component that remains consistently beneficial across all eight benchmark families and network capacities, justifying its role as the regime-agnostic core of \method{}.
Notably, the aggregate gains are often obtained by \emph{removing} conservative components rather than adding new ones, reframing scalable off-policy RL as a regime-matching problem rather than a stabilizer-stacking one.

\vspace{-0.2cm}
\paragraph{Limitations and Future Work.}
Our current prescription selects between \method{}-L and \method{}-A offline based on the target regime; deployments that span regimes---for example, pretraining under narrow data followed by parallel fine-tuning---would benefit from an \emph{online regime-adaptive} variant that monitors replay coverage or value-extrapolation signals and continuously modulates normalization strength and critic multiplicity.
More broadly, our analysis is grounded in the FlashSAC backbone and focuses on three axes; a natural next step is to examine whether other canonical stabilizers, e.g., entropy weighting, target-network delay, gradient clipping, or replay-ratio schedules, share the same regime-dependent behavior.

These findings suggest that scalable RL requires data-regime-aware algorithm design.
Classical stabilizers such as normalization and double-Q targets are valuable when data are scarce, but may impose unnecessary bias or computation when parallel simulation already provides broad replay coverage.
A promising direction is to adapt these mechanisms online, retaining conservative constraints when uncertainty or coverage is poor and relaxing them when exploitation of abundant replay data becomes the main bottleneck.
\bibliographystyle{assets/plainnat}
\bibliography{references}

\begin{thebibliography}{32}
\providecommand{\natexlab}[1]{#1}
\providecommand{\url}[1]{\texttt{#1}}
\expandafter\ifx\csname urlstyle\endcsname\relax
  \providecommand{\doi}[1]{doi: #1}\else
  \providecommand{\doi}{doi: \begingroup \urlstyle{rm}\Url}\fi

\bibitem[Ba et~al.(2016)Ba, Kiros, and Hinton]{ba2016layer}
Jimmy~Lei Ba, Jamie~Ryan Kiros, and Geoffrey~E. Hinton.
\newblock Layer normalization.
\newblock \emph{arXiv preprint arXiv:1607.06450}, 2016.

\bibitem[Bartlett et~al.(2017)Bartlett, Foster, and
  Telgarsky]{bartlett2017spectrally}
Peter~L. Bartlett, Dylan~J. Foster, and Matus~J. Telgarsky.
\newblock Spectrally-normalized margin bounds for neural networks.
\newblock In \emph{Advances in Neural Information Processing Systems},
  volume~30, 2017.

\bibitem[Caggiano et~al.(2022)Caggiano, Wang, Durandau, Sartori, and
  Kumar]{caggiano2022myosuite}
Vittorio Caggiano, Huawei Wang, Guillaume Durandau, Massimo Sartori, and Vikash
  Kumar.
\newblock {MyoSuite}: A contact-rich simulation suite for musculoskeletal motor
  control.
\newblock In \emph{Proceedings of The 4th Annual Learning for Dynamics and
  Control Conference}, volume 168 of \emph{Proceedings of Machine Learning
  Research}, pages 492--507. PMLR, 2022.

\bibitem[Chen et~al.(2021)Chen, Wang, Zhou, and Ross]{chen2021redq}
Xinyue Chen, Che Wang, Zijian Zhou, and Keith~W. Ross.
\newblock Randomized ensembled double q-learning: Learning fast without a
  model.
\newblock In \emph{International Conference on Learning Representations}, 2021.

\bibitem[Fujimoto and Gu(2023)]{fujimoto2023td7}
Scott Fujimoto and Shixiang~Shane Gu.
\newblock {TD7}: A high-performance algorithm for continuous control, 2023.
\newblock \url{https://arxiv.org/abs/2306.02451}.

\bibitem[Fujimoto et~al.(2018)Fujimoto, van Hoof, and Meger]{fujimoto2018td3}
Scott Fujimoto, Herke van Hoof, and David Meger.
\newblock Addressing function approximation error in actor-critic methods.
\newblock In \emph{Proceedings of the 35th International Conference on Machine
  Learning}, volume~80 of \emph{Proceedings of Machine Learning Research},
  pages 1587--1596. PMLR, 2018.

\bibitem[Haarnoja et~al.(2018{\natexlab{a}})Haarnoja, Zhou, Abbeel, and
  Levine]{haarnoja2018sac}
Tuomas Haarnoja, Aurick Zhou, Pieter Abbeel, and Sergey Levine.
\newblock Soft actor-critic: Off-policy maximum entropy deep reinforcement
  learning with a stochastic actor.
\newblock In \emph{Proceedings of the 35th International Conference on Machine
  Learning}, volume~80 of \emph{Proceedings of Machine Learning Research},
  pages 1861--1870. PMLR, 2018{\natexlab{a}}.

\bibitem[Haarnoja et~al.(2018{\natexlab{b}})Haarnoja, Zhou, Hartikainen,
  Tucker, Ha, Tan, Kumar, Zhu, Gupta, Abbeel, and
  Levine]{haarnoja2018applications}
Tuomas Haarnoja, Aurick Zhou, Kristian Hartikainen, George Tucker, Sehoon Ha,
  Jie Tan, Vikash Kumar, Henry Zhu, Abhishek Gupta, Pieter Abbeel, and Sergey
  Levine.
\newblock Soft actor-critic algorithms and applications.
\newblock \emph{arXiv preprint arXiv:1812.05905}, 2018{\natexlab{b}}.

\bibitem[Hansen et~al.(2024)Hansen, Su, and Wang]{hansen2024tdmpc2}
Nicklas Hansen, Hao Su, and Xiaolong Wang.
\newblock {TD-MPC2}: Scalable, robust world models for continuous control.
\newblock \emph{arXiv preprint arXiv:2310.16828}, 2024.

\bibitem[Ioffe and Szegedy(2015)]{ioffe2015batch}
Sergey Ioffe and Christian Szegedy.
\newblock Batch normalization: Accelerating deep network training by reducing
  internal covariate shift.
\newblock In \emph{Proceedings of the 32nd International Conference on Machine
  Learning}, volume~37 of \emph{Proceedings of Machine Learning Research},
  pages 448--456. PMLR, 2015.

\bibitem[Kim et~al.(2026)Kim, Lee, Park, Kim, Nahendra, Seno, Min, Palenicek,
  Vogt, Kragic, Peters, Choo, and Lee]{kim2026flashsac}
Donghu Kim, Youngdo Lee, Minho Park, Kinam Kim, I~Made~Aswin Nahendra, Takuma
  Seno, Sehee Min, Daniel Palenicek, Florian Vogt, Danica Kragic, Jan Peters,
  Jaegul Choo, and Hojoon Lee.
\newblock {FlashSAC}: Fast and stable off-policy reinforcement learning for
  high-dimensional robot control.
\newblock In \emph{Robotics: Science and Systems}, 2026.
\newblock \url{https://arxiv.org/abs/2604.04539}.

\bibitem[Lee et~al.(2025)Lee, Lee, Seno, Kim, Stone, and
  Choo]{lee2025hyperspherical}
Hojoon Lee, Youngdo Lee, Takuma Seno, Donghu Kim, Peter Stone, and Jaegul Choo.
\newblock Hyperspherical normalization for scalable deep reinforcement
  learning.
\newblock In \emph{International Conference on Machine Learning}, 2025.
\newblock \url{https://arxiv.org/abs/2502.15280}.

\bibitem[Ma et~al.(2023)Ma, Tang, Li, and Meng]{ma2023reining}
Yi~Ma, Hongyao Tang, Dong Li, and Zhaopeng Meng.
\newblock Reining generalization in offline reinforcement learning via
  representation distinction.
\newblock In \emph{Advances in Neural Information Processing Systems},
  volume~36, pages 40773--40785, 2023.

\bibitem[Mittal et~al.(2025)Mittal, Roth, Tigue, Richard, Zhang, Du,
  Serrano-Muñoz, Yao, Zurbrügg, Rudin, Wawrzyniak, Rakhsha, Denzler, Heiden,
  Borovicka, Ahmed, Akinola, Anwar, Carlson, Feng, Garg, Gasoto, Gulich, Guo,
  Gussert, Hansen, Kulkarni, Li, Liu, Makoviychuk, Malczyk, Mazhar, Moghani,
  Murali, Noseworthy, Poddubny, Ratliff, Rehberg, Schwarke, Singh, Smith, Tang,
  Thaker, Trepte, Wyk, Yu, Millane, Ramasamy, Steiner, Subramanian, Volk, Chen,
  Jawale, Kuruttukulam, Lin, Mandlekar, Patzwaldt, Welsh, Zhao, Anes, Lafleche,
  Moënne-Loccoz, Park, Stepinski, Gelder, Amevor, Carius, Chang, Chen,
  de~Heras~Ciechomski, Daviet, Mohajerani, von Muralt, Reutskyy, Sauter,
  Schirm, Shi, Terdiman, Vilella, Widmer, Yeoman, Chen, Grizan, Li, Li, Smith,
  Wiltz, Alexis, Chang, Chu, Fan, Farshidian, Handa, Huang, Hutter, Narang,
  Pouya, Sheng, Zhu, Macklin, Moravanszky, Reist, Guo, Hoeller, and
  State]{mittal2025isaaclab}
Mayank Mittal, Pascal Roth, James Tigue, Antoine Richard, Octi Zhang, Peter Du,
  Antonio Serrano-Muñoz, Xinjie Yao, René Zurbrügg, Nikita Rudin, Lukasz
  Wawrzyniak, Milad Rakhsha, Alain Denzler, Eric Heiden, Ales Borovicka, Ossama
  Ahmed, Iretiayo Akinola, Abrar Anwar, Mark~T. Carlson, Ji~Yuan Feng, Animesh
  Garg, Renato Gasoto, Lionel Gulich, Yijie Guo, M.~Gussert, Alex Hansen, Mihir
  Kulkarni, Chenran Li, Wei Liu, Viktor Makoviychuk, Grzegorz Malczyk, Hammad
  Mazhar, Masoud Moghani, Adithyavairavan Murali, Michael Noseworthy, Alexander
  Poddubny, Nathan Ratliff, Welf Rehberg, Clemens Schwarke, Ritvik Singh,
  James~Latham Smith, Bingjie Tang, Ruchik Thaker, Matthew Trepte, Karl~Van
  Wyk, Fangzhou Yu, Alex Millane, Vikram Ramasamy, Remo Steiner, Sangeeta
  Subramanian, Clemens Volk, CY~Chen, Neel Jawale, Ashwin~Varghese
  Kuruttukulam, Michael~A. Lin, Ajay Mandlekar, Karsten Patzwaldt, John Welsh,
  Huihua Zhao, Fatima Anes, Jean-Francois Lafleche, Nicolas Moënne-Loccoz,
  Soowan Park, Rob Stepinski, Dirk~Van Gelder, Chris Amevor, Jan Carius,
  Jumyung Chang, Anka~He Chen, Pablo de~Heras~Ciechomski, Gilles Daviet,
  Mohammad Mohajerani, Julia von Muralt, Viktor Reutskyy, Michael Sauter, Simon
  Schirm, Eric~L. Shi, Pierre Terdiman, Kenny Vilella, Tobias Widmer, Gordon
  Yeoman, Tiffany Chen, Sergey Grizan, Cathy Li, Lotus Li, Connor Smith, Rafael
  Wiltz, Kostas Alexis, Yan Chang, David Chu, Linxi~{Jim} Fan, Farbod
  Farshidian, Ankur Handa, Spencer Huang, Marco Hutter, Yashraj Narang, Soha
  Pouya, Shiwei Sheng, Yuke Zhu, Miles Macklin, Adam Moravanszky, Philipp
  Reist, Yunrong Guo, David Hoeller, and Gavriel State.
\newblock Isaac lab: A gpu-accelerated simulation framework for multi-modal
  robot learning.
\newblock \emph{arXiv preprint arXiv:2511.04831}, 2025.
\newblock \url{https://arxiv.org/abs/2511.04831}.

\bibitem[Miyato et~al.(2018)Miyato, Kataoka, Koyama, and
  Yoshida]{miyato2018spectral}
Takeru Miyato, Toshiki Kataoka, Masanori Koyama, and Yuichi Yoshida.
\newblock Spectral normalization for generative adversarial networks.
\newblock In \emph{International Conference on Learning Representations}, 2018.

\bibitem[Nikishin et~al.(2022)Nikishin, Schwarzer, D'Oro, Bacon, and
  Courville]{nikishin2022primacy}
Evgenii Nikishin, Max Schwarzer, Pierluca D'Oro, Pierre-Luc Bacon, and Aaron
  Courville.
\newblock The primacy bias in deep reinforcement learning.
\newblock In \emph{Proceedings of the 39th International Conference on Machine
  Learning}, Proceedings of Machine Learning Research. PMLR, 2022.

\bibitem[Osband and Van~Roy(2014)]{osband2014model}
Ian Osband and Benjamin Van~Roy.
\newblock Model-based reinforcement learning and the eluder dimension.
\newblock In \emph{Advances in Neural Information Processing Systems},
  volume~27, 2014.

\bibitem[Palenicek et~al.(2026)Palenicek, Vogt, Watson, Posner, and
  Peters]{palenicek2026xqc}
Daniel Palenicek, Florian Vogt, Joe Watson, Ingmar Posner, and Jan Peters.
\newblock {XQC}: Well-conditioned optimization accelerates deep reinforcement
  learning.
\newblock In \emph{International Conference on Learning Representations}, 2026.
\newblock \url{https://arxiv.org/abs/2509.25174}.

\bibitem[Russo and Van~Roy(2013)]{russo2013eluder}
Daniel Russo and Benjamin Van~Roy.
\newblock Eluder dimension and the sample complexity of optimistic exploration.
\newblock In \emph{Advances in Neural Information Processing Systems},
  volume~26, 2013.

\bibitem[Salimans and Kingma(2016)]{salimans2016weight}
Tim Salimans and Diederik~P. Kingma.
\newblock Weight normalization: A simple reparameterization to accelerate
  training of deep neural networks.
\newblock In \emph{Advances in Neural Information Processing Systems},
  volume~29, 2016.

\bibitem[Schaul et~al.(2016)Schaul, Quan, Antonoglou, and
  Silver]{schaul2016prioritized}
Tom Schaul, John Quan, Ioannis Antonoglou, and David Silver.
\newblock Prioritized experience replay.
\newblock In \emph{International Conference on Learning Representations}, 2016.

\bibitem[Sferrazza et~al.(2024)Sferrazza, Huang, Lin, Lee, and
  Abbeel]{sferrazza2024humanoidbench}
Carmelo Sferrazza, Dun-Ming Huang, Xingyu Lin, Youngwoon Lee, and Pieter
  Abbeel.
\newblock {HumanoidBench}: Simulated humanoid benchmark for whole-body
  locomotion and manipulation.
\newblock \emph{arXiv preprint arXiv:2403.10506}, 2024.

\bibitem[Sokar et~al.(2023)Sokar, Agarwal, Castro, and Evci]{sokar2023redo}
Ghada Sokar, Rishabh Agarwal, Pablo~Samuel Castro, and Utku Evci.
\newblock The dormant neuron phenomenon in deep reinforcement learning.
\newblock In \emph{Proceedings of the 40th International Conference on Machine
  Learning}, Proceedings of Machine Learning Research. PMLR, 2023.

\bibitem[Tang and Berseth(2024)]{tang2024chain}
Hongyao Tang and Glen Berseth.
\newblock Improving deep reinforcement learning by reducing the chain effect of
  value and policy churn.
\newblock In \emph{Advances in Neural Information Processing Systems},
  volume~37, 2024.

\bibitem[Tang et~al.(2025)Tang, Obando{-}Ceron, Castro, Courville, and
  Berseth]{Tang25CCHAIN}
Hongyao Tang, Johan~S. Obando{-}Ceron, Pablo~Samuel Castro, Aaron~C. Courville,
  and Glen Berseth.
\newblock Mitigating plasticity loss in continual reinforcement learning by
  reducing churn.
\newblock In \emph{ICML}, 2025.

\bibitem[Tao et~al.(2024)Tao, Xiang, Shukla, Qin, Hinrichsen, Yuan, Bao, Lin,
  Liu, kai Chan, Gao, Li, Mu, Xiao, Gurha, Huang, Calandra, Chen, Luo, and
  Su]{taomaniskill3}
Stone Tao, Fanbo Xiang, Arth Shukla, Yuzhe Qin, Xander Hinrichsen, Xiaodi Yuan,
  Chen Bao, Xinsong Lin, Yulin Liu, Tse kai Chan, Yuan Gao, Xuanlin Li,
  Tongzhou Mu, Nan Xiao, Arnav Gurha, Zhiao Huang, Roberto Calandra, Rui Chen,
  Shan Luo, and Hao Su.
\newblock Maniskill3: Gpu parallelized robotics simulation and rendering for
  generalizable embodied ai.
\newblock \emph{arXiv preprint arXiv:2410.00425}, 2024.

\bibitem[Tassa et~al.(2018)Tassa, Doron, Muldal, Erez, Li, de~Las~Casas,
  Budden, Abdolmaleki, Merel, Lefrancq, Lillicrap, and
  Riedmiller]{tassa2018deepmind}
Yuval Tassa, Yotam Doron, Alistair Muldal, Tom Erez, Yazhe Li, Diego
  de~Las~Casas, David Budden, Abbas Abdolmaleki, Josh Merel, Andrew Lefrancq,
  Timothy Lillicrap, and Martin Riedmiller.
\newblock Deepmind control suite.
\newblock \emph{arXiv preprint arXiv:1801.00690}, 2018.

\bibitem[Todorov et~al.(2012)Todorov, Erez, and Tassa]{todorov2012mujoco}
Emanuel Todorov, Tom Erez, and Yuval Tassa.
\newblock {MuJoCo}: A physics engine for model-based control.
\newblock In \emph{2012 IEEE/RSJ International Conference on Intelligent Robots
  and Systems}, pages 5026--5033. IEEE, 2012.

\bibitem[Wu et~al.(2026)Wu, Tang, Ma, Liu, Zheng, and Hao]{wu2026rank}
Zihao Wu, Hongyao Tang, Yi~Ma, Jiashun Liu, Yan Zheng, and Jianye Hao.
\newblock The rank and gradient lost in non-stationarity: Sample weight decay
  for mitigating plasticity loss in reinforcement learning.
\newblock In \emph{International Conference on Learning Representations}, 2026.
\newblock \url{https://openreview.net/forum?id=5DpzzTPnJZ}.

\bibitem[Zakka et~al.(2025)Zakka, Tabanpour, Liao, Haiderbhai, Holt, Luo,
  Allshire, Frey, Sreenath, Kahrs, Sferrazza, Tassa, and
  Abbeel]{zakka2025mujoco}
Kevin Zakka, Baruch Tabanpour, Qiayuan Liao, Mustafa Haiderbhai, Samuel Holt,
  Jing~Yuan Luo, Arthur Allshire, Erik Frey, Koushil Sreenath, Lueder~A. Kahrs,
  Carmelo Sferrazza, Yuval Tassa, and Pieter Abbeel.
\newblock {MuJoCo Playground}, 2025.
\newblock \url{https://arxiv.org/abs/2502.08844}.

\bibitem[Zakka et~al.(2026)Zakka, Liao, Yi, Le~Lay, Sreenath, and
  Abbeel]{zakka2026mjlab}
Kevin Zakka, Qiayuan Liao, Brent Yi, Louis Le~Lay, Koushil Sreenath, and Pieter
  Abbeel.
\newblock {mjlab}: A lightweight framework for gpu-accelerated robot learning,
  2026.
\newblock \url{https://arxiv.org/abs/2601.22074}.

\bibitem[Zhang and Sennrich(2019)]{zhang2019rmsnorm}
Biao Zhang and Rico Sennrich.
\newblock Root mean square layer normalization.
\newblock In \emph{Advances in Neural Information Processing Systems},
  volume~32, 2019.

\end{thebibliography}

\clearpage
\begin{center}
    {\Huge\bfseries\textcolor{appendixpurple}{Appendix Contents}}
\end{center}
\vspace{1.0em}

\begin{tcolorbox}[
    colback=appendixlightpurple,
    colframe=appendixpurple,
    boxrule=0.8pt,
    arc=2pt,
    left=1.1em,
    right=1.1em,
    top=1.0em,
    bottom=1.0em
]
\begingroup
\renewcommand{\arraystretch}{1.55}
\setlength{\tabcolsep}{0pt}
\begin{tabularx}{\textwidth}{@{}p{3.2em}@{\hspace{1.4em}}Xr@{}}
\appentry{A}{The Use of Large Language Models}{app:llm}
\appentry{B}{Impact Statement}{app:impact}
\appentry{C}{Wall-Clock Time}{app:wall-time}
\appsubentry{C.1}{Update}{app:wall-time-update}
\appsubentry{C.2}{SWD Sampling}{app:wall-time-swd}
\appsubentry{C.3}{A800 End-to-End Wall Time}{app:wall-time-a800}
\appsubentry{C.4}{GPU-Parallel Learning Curves by Wall Time}{app:gpu-wall-time-curves}
\appentry{D}{Experimental Details}{app:experiments}
\appsubentry{D.1}{SWD Implementations}{app:swd}
\appsubentry{D.2}{Hyperparameters}{app:hyperparameter}
\appsubentry{D.3}{Network Architecture}{app:network}
\appentry{E}{Environment Details}{app:environment}
\appentry{F}{Full Results}{app:results}
\appsubentry{F.1}{Gym--MuJoCo}{app:results-mujoco}
\appsubentry{F.2}{DMC}{app:results-dmc}
\appsubentry{F.3}{MyoSuite}{app:results-myosuite}
\appsubentry{F.4}{HumanoidBench}{app:results-humanoidbench}
\appsubentry{F.5}{MuJoCo Playground}{app:results-playground}
\appsubentry{F.6}{IsaacLab}{app:results-isaaclab}
\appsubentry{F.7}{MJLab}{app:results-mjlab}
\appsubentry{F.8}{ManiSkill}{app:results-maniskill}
\appsubentry{F.9}{Ablation Learning Curves}{app:results-ablation}
\end{tabularx}
\endgroup
\end{tcolorbox}

\clearpage
\beginappendix
\captionsetup[table]{position=top,skip=6pt,labelfont=bf,labelsep=space}
\newcolumntype{L}{>{\raggedright\arraybackslash}X}
\newcolumntype{C}{>{\centering\arraybackslash}X}
\newcolumntype{Y}{>{\ttfamily\raggedright\arraybackslash}X}
\newcolumntype{G}[1]{>{\centering\arraybackslash}p{#1}}
\newcolumntype{P}[1]{>{\raggedright\arraybackslash}p{#1}}
\newcommand{\envtablesetup}{%
    \small
    \setlength{\tabcolsep}{12pt}%
    \renewcommand{\arraystretch}{1.08}%
}
\newcommand{\tablehead}[1]{\textnormal{\textbf{#1}}}
\newcommand{\tablegroup}[3]{\multirow{#1}{*}{\makebox[#2][c]{\textbf{#3}}}}
\newcommand{\envhead}[1]{\tablehead{#1}}

\section{The Use of Large Language Models}
\label{app:llm}
LLMs were used only for language polishing and grammar correction. All ideas, method design, analysis, figures, and experimental results were produced by the authors.

\section{Impact Statement}
\label{app:impact}
This work studies reinforcement-learning algorithms for simulated control. It does not introduce new deployment systems or human-subject data. Potential societal effects are therefore indirect and mainly relate to future uses of improved control algorithms.

\section{Wall-Clock Time}
\label{app:wall-time}
\subsection{Update}
\label{app:wall-time-update}
We measure the wall-clock cost of one complete SAC update after reimplementing the FlashSAC backbone in JAX with Flax NNX. This benchmark isolates learner computation from environment stepping, replay-buffer sampling, logging, and evaluation. We use synthetic batches on an NVIDIA GeForce RTX 5060 Ti with observation dimension 128, action dimension 12, and batch size 2048. Each result is measured after JIT or \texttt{torch.compile} warmup, using 20 warmup repetitions and 80 measured repetitions per round over three alternating rounds. We report the median over the three round-level means. For PyTorch AMP, we use the FlashSAC implementation with CUDA autocast to \texttt{float16} and \texttt{GradScaler}.

\begin{table}[H]
    \centering
    \caption{\textbf{SAC update micro-benchmark.} Lower is better.}
    \label{tab:wall_clock_microbench}
    \envtablesetup
    \begin{tabularx}{0.76\textwidth}{@{}Lc@{}}
        \toprule
        \tablehead{Implementation} & \tablehead{Full Update} \\
        \midrule
        JAX/NNX bfloat16 & 7.895 ms \\
        JAX/NNX float32 & 13.244 ms \\
        PyTorch FlashSAC, \texttt{torch.compile}, AMP on & 13.922 ms \\
        PyTorch FlashSAC, \texttt{torch.compile}, AMP off & 12.293 ms \\
        \bottomrule
    \end{tabularx}
\end{table}

The JAX/NNX bfloat16 update takes 7.895 ms, which is $1.68\times$ faster than JAX/NNX float32, $1.76\times$ faster than compiled PyTorch FlashSAC with AMP, and $1.56\times$ faster than compiled PyTorch FlashSAC without AMP under the same synthetic-batch setting. In the compiled PyTorch implementation, AMP accelerates the critic update in isolation, but the complete update is slightly faster without AMP because the full step also includes the actor update, temperature update, normalization, and scaler/autocast overhead. These measurements suggest that the main computational benefit of the JAX/NNX implementation comes from the compiled update path, while bfloat16 further improves update throughput.

\subsection{SWD Sampling}
\label{app:wall-time-swd}
We also measure the wall-clock overhead introduced by SWD replay sampling. The benchmark uses the same NVIDIA GeForce RTX 5060 Ti and the same synthetic transition shape as above, with a replay capacity of $10^6$, batch size 2048, observation dimension 128, and action dimension 12. We compare uniform replay, exact SWD sampling, and the bucketed approximate SWD sampler with 2000 buckets. The SWD decay horizon is $80{,}000$ and $w_{\min}=0.1$. Each result measures only \texttt{JaxBuffer.sample} after JIT warmup, using 80 warmup repetitions and 300 measured repetitions per round over nine alternating rounds. We report the median over the nine round-level means.

\begin{table}[H]
    \centering
    \caption{\textbf{Replay sampling micro-benchmark.} Lower is better.}
    \label{tab:swd_sampling_wall_clock}
    \envtablesetup
    \begin{tabularx}{0.78\textwidth}{@{}Lcc@{}}
        \toprule
        \tablehead{Sampler} & \tablehead{CPU Sample} & \tablehead{GPU Sample} \\
        \midrule
        Uniform replay & 0.710 ms & 0.267 ms \\
        Exact SWD & 3.263 ms & 1.751 ms \\
        Approximate SWD, 2000 buckets & 0.959 ms & 1.187 ms \\
        \bottomrule
    \end{tabularx}
\end{table}

On CPU, exact SWD is $4.59\times$ slower than uniform replay because it constructs and normalizes a probability vector over the full million-transition buffer at each sample call. The bucketed approximation avoids this full scan and reduces sampling time from 3.263 ms to 0.959 ms, although it remains $1.35\times$ slower than uniform replay. On GPU, exact SWD costs 1.751 ms per sample and is $6.56\times$ slower than uniform replay, while the bucketed approximation reduces this cost to 1.187 ms. In absolute terms, however, the exact GPU sampling overhead is still small relative to learner computation: exact SWD sampling plus a full JAX/NNX bfloat16 update takes about 9.646 ms, which remains faster than compiled PyTorch FlashSAC with AMP (13.922 ms) and without AMP (12.293 ms). These results support using the approximate sampler for CPU-scale replay buffers and exact SWD for GPU-parallel experiments, where preserving the exact sampling distribution is affordable and does not erase the throughput advantage of the JAX/NNX bfloat16 implementation.

\vspace{-0.25cm}
\subsection{A800 End-to-End Wall Time}
\label{app:wall-time-a800}
We measure end-to-end wall-clock learning on NVIDIA A800 GPUs. This includes environment stepping, replay sampling, learner updates, logging, and evaluation. We report four completed IsaacLab environments and average \method{} w Norm OFF over five seeds.

\begin{figure}[H]
    \vspace{-0.35cm}
    \centering
    \includegraphics[width=\textwidth]{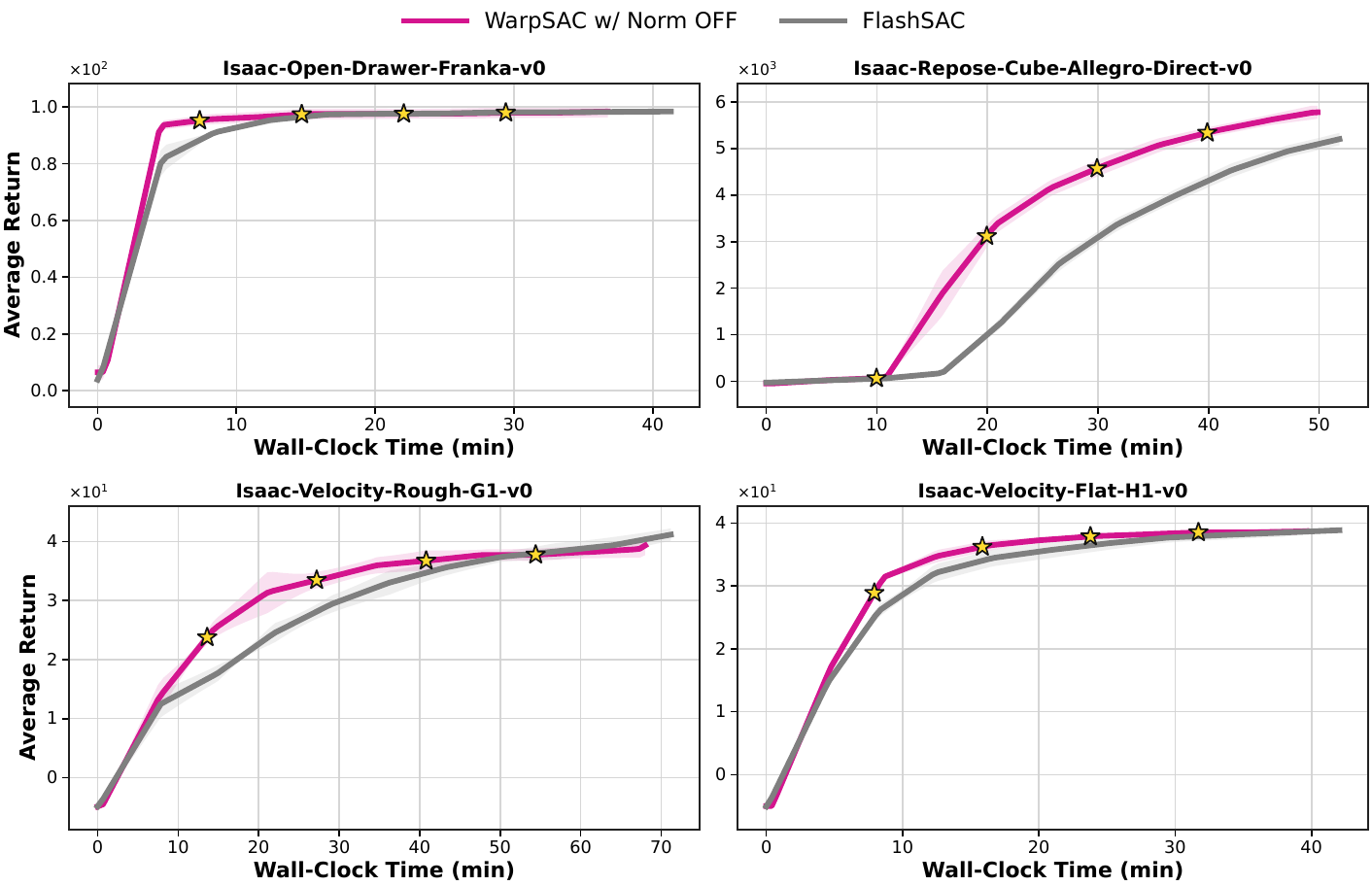}
    \caption{Actual A800 return--wall-time curves on four completed IsaacLab environments. Curves report average return versus elapsed wall-clock time.}
    \label{fig:a800_wall_time}
    \vspace{-0.35cm}
\end{figure}

Figure~\ref{fig:a800_wall_time} shows that \method{} w Norm OFF reaches the 50.0M-step endpoint at a similar or slightly faster wall-clock speed than FlashSAC on all four selected environments. The final wall-clock speedups are $1.11\times$ on Isaac-Open-Drawer-Franka-v0, $1.03\times$ on Isaac-Repose-Cube-Allegro-Direct-v0, $1.04\times$ on Isaac-Velocity-Rough-G1-v0, and $1.05\times$ on Isaac-Velocity-Flat-H1-v0. Final returns are comparable on Isaac-Open-Drawer-Franka-v0, Isaac-Velocity-Rough-G1-v0, and Isaac-Velocity-Flat-H1-v0, while \method{} w Norm OFF obtains a higher final return on Isaac-Repose-Cube-Allegro-Direct-v0.

These real-machine measurements indicate that the JAX/NNX implementation does not introduce a meaningful end-to-end slowdown relative to FlashSAC in GPU-parallel IsaacLab training. We therefore use the measured \method{} wall-time traces as a practical speed proxy only for domains whose FlashSAC logs contain sample-efficiency curves but not full wall-clock traces. In those cases, the compute-efficiency figures below plot FlashSAC on the wall-time axis of the corresponding \method{} Norm ON, num-Q $=2$ run, which keeps the speed assumption matched to the closest two-critic baseline and avoids attributing additional speed differences to FlashSAC.

\subsection{GPU-Parallel Learning Curves by Wall Time}
\label{app:gpu-wall-time-curves}

\begin{figure}[H]
    \centering
    \vspace{-0.2cm}
    \includegraphics[width=\linewidth]{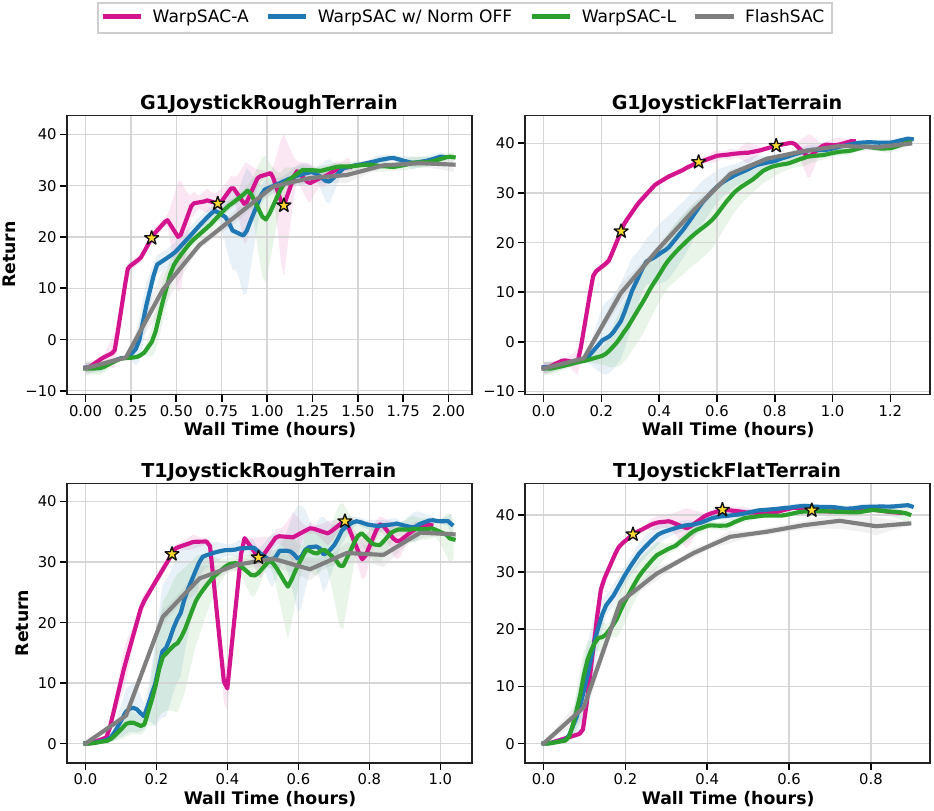}
    \vspace{-0.2cm}
    \caption{\textbf{MuJoCo Playground learning curves by wall-clock time.}
    Curves report mean return with standard deviation across five seeds. FlashSAC is plotted using the same wall-time axis as the corresponding \method{} Norm ON, num-Q $=2$ run.}
    \label{fig:wall_time_playground}
    \vspace{-0.35cm}
\end{figure}

\begin{figure}[H]
    \centering
    \vspace{-0.2cm}
    \includegraphics[width=\linewidth]{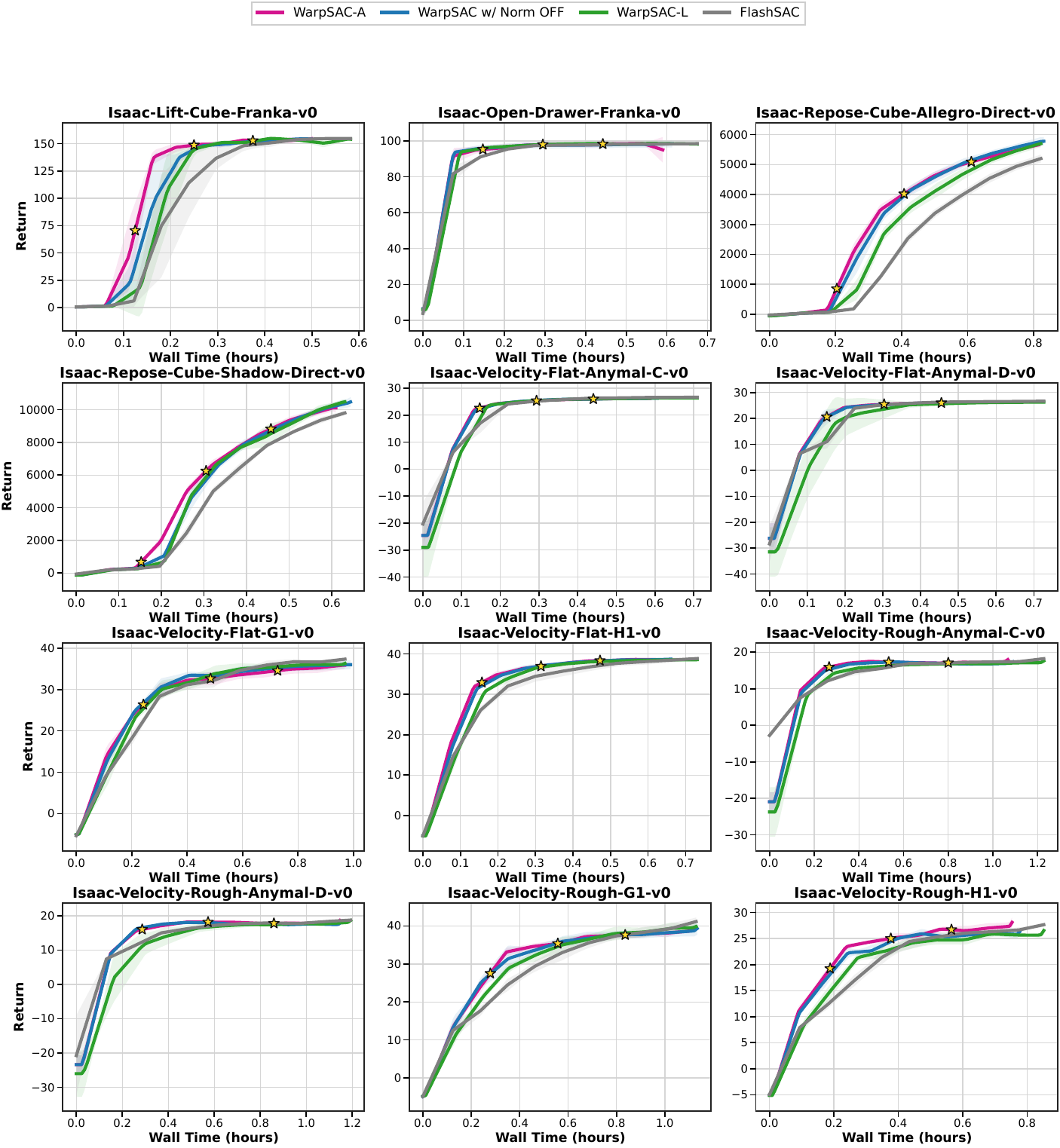}
    \vspace{-0.2cm}
    \caption{\textbf{IsaacLab learning curves by wall-clock time.}
    Curves report mean return with standard deviation across five seeds for the evaluated GPU-parallel IsaacLab tasks. FlashSAC is plotted using the same wall-time axis as the corresponding \method{} Norm ON, num-Q $=2$ run.}
    \label{fig:wall_time_isaaclab}
    \vspace{-0.35cm}
\end{figure}

\begin{figure}[H]
    \centering
    \vspace{-0.2cm}
    \includegraphics[width=\linewidth]{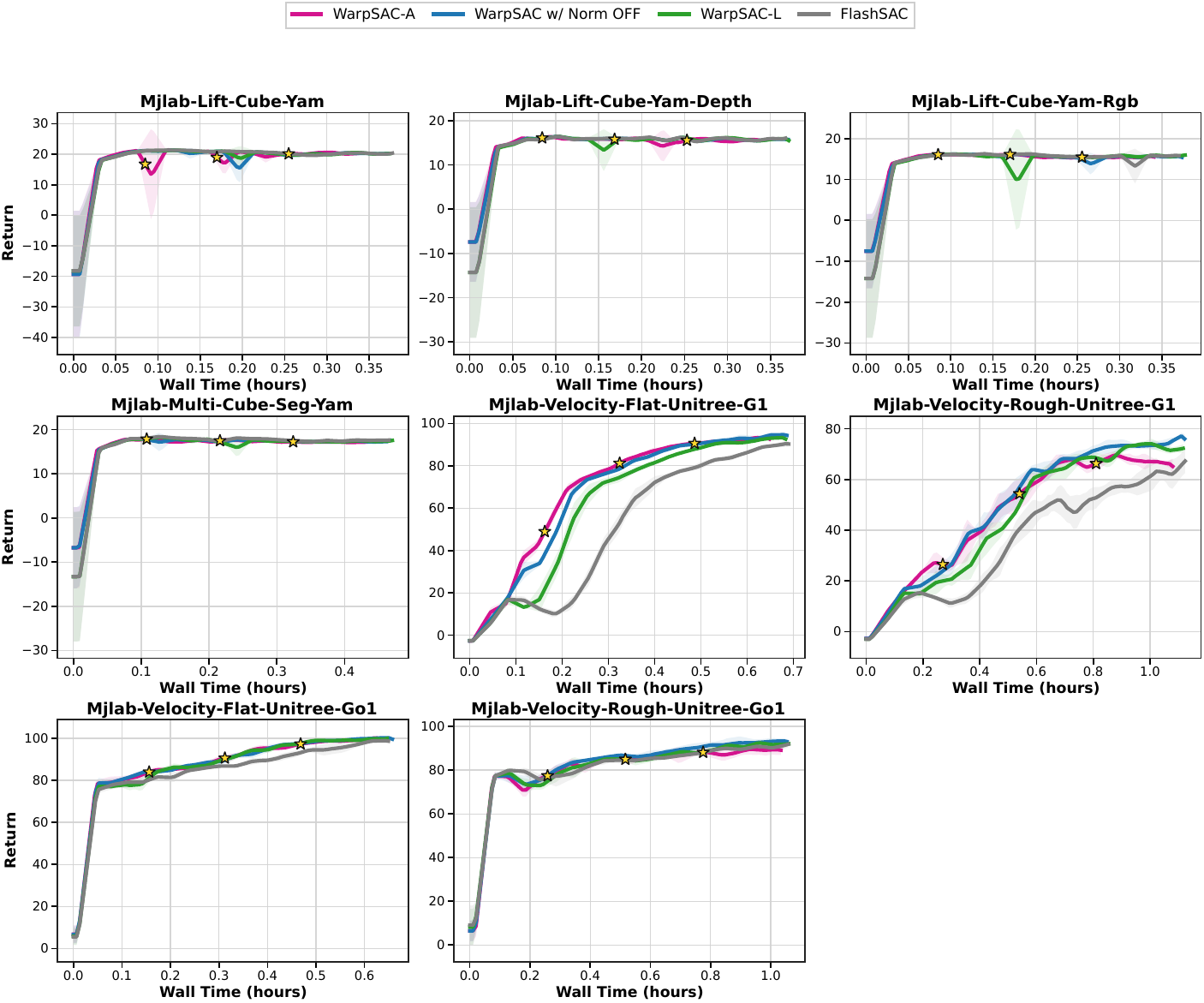}
    \vspace{-0.2cm}
    \caption{\textbf{MJLab learning curves by wall-clock time.}
    Curves report mean return with standard deviation across five seeds for Unitree locomotion and Yam manipulation tasks. All methods, including FlashSAC, use their own logged wall-clock traces.}
    \label{fig:wall_time_mjlab}
    \vspace{-0.35cm}
\end{figure}

\begin{figure}[H]
    \centering
    \vspace{-0.2cm}
    \includegraphics[width=\linewidth]{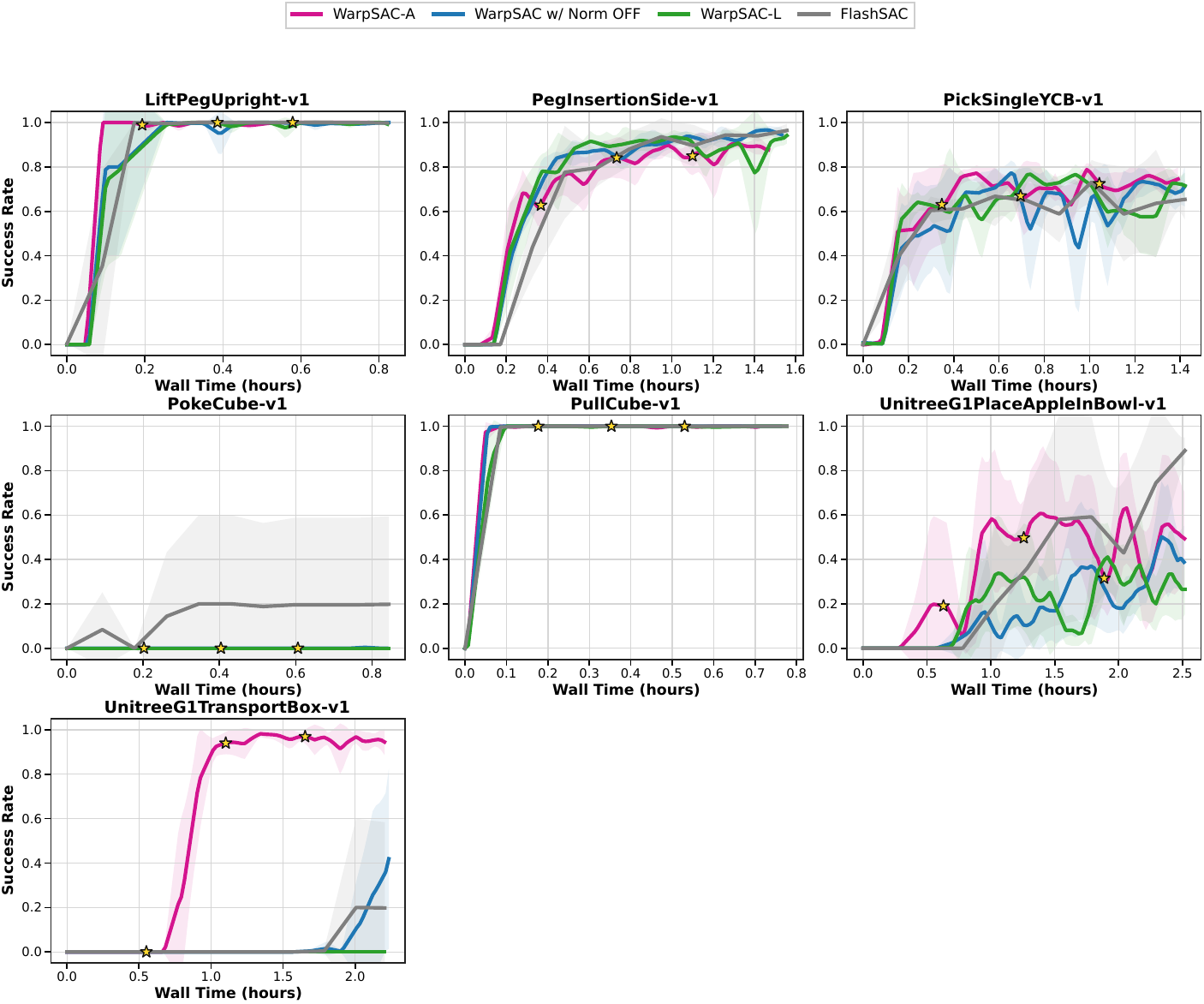}
    \vspace{-0.2cm}
    \caption{\textbf{ManiSkill learning curves by wall-clock time.}
    Curves report mean success-once rate with standard deviation across five seeds. FlashSAC is plotted using the same wall-time axis as the corresponding \method{} Norm ON, num-Q $=2$ run.}
    \label{fig:wall_time_maniskill}
    \vspace{-0.35cm}
\end{figure}

Figures~\ref{fig:wall_time_playground}--\ref{fig:wall_time_maniskill} report GPU-parallel learning curves against wall-clock time for MuJoCo Playground, IsaacLab, MJLab, and ManiSkill. For MJLab, every method is plotted against its own logged wall-clock trace, including FlashSAC. For the other three domains, the FlashSAC sample-efficiency curves use the wall-time axis of the corresponding \method{} Norm ON, num-Q $=2$ run, following the speed-proxy convention motivated above.

\section{Experimental Details}
\label{app:experiments}

\subsection{SWD Implementations}
\label{app:swd}

SWD is implemented in the replay buffer. Each transition receives an insertion timestamp, and sampling weights are computed from transition age. Setting the decay horizon to zero recovers uniform replay, so FlashSAC and SWD variants share the same buffer implementation.

We use bucketed approximate sampling only for CPU-scale replay buffers: DMC hard tasks, MuJoCo, MyoSuite, and HumanoidBench. These runs store replay on the host with capacity $10^6$. GPU-parallel domains, including MuJoCo Playground, MJLab, ManiSkill, and IsaacLab, use exact SWD because replay tensors are stored on GPU and categorical sampling is CUDA-accelerated through JAX.

The wall-clock micro-benchmark in Appendix~\ref{app:wall-time-swd} supports this choice. With a $10^6$-transition replay buffer and batch size 2048, bucketed CPU sampling reduces exact SWD sampling from 3.263 ms to 0.959 ms. On GPU, exact SWD sampling costs 1.751 ms, and exact SWD sampling plus a full JAX/NNX bfloat16 update remains faster than compiled PyTorch FlashSAC under the same synthetic-batch setting.

For a transition inserted at time $t_i$ and sampled at replay time $t$, the age is $A_t(i)=t-t_i$. In the main experiments we use a recent-sample bias,
\[
    w_t(i)=\max\left(w_{\min}, 1-\frac{A_t(i)}{T_{\mathrm{decay}}}\right),
\]
where $w_{\min}=0.1$ in all reported runs. The resulting sampling probability is
\[
    p_t(i)=\frac{w_t(i)}{\sum_j w_t(j)}.
\]
The implementation also supports the opposite bias direction through a negative decay horizon, but this option is not used in the reported experiments.

\paragraph{Exact Numpy implementation.}
The following code is the core Numpy implementation used for exact age-biased sampling. It first constructs a probability distribution over all valid replay entries and then samples minibatch indices with \texttt{np.random.choice}.

\begin{lstlisting}[style=pythonappendix,caption={Exact Numpy implementation of Sample Weight Decay.},label={lst:swd_exact_numpy}]
def _sample_indices(self, batch_size: int) -> np.ndarray:
    if self.linear_decay_step == 0:
        return np.random.randint(0, self.size, size=batch_size)

    if self.use_approximate_sampling:
        return self._sample_indices_with_approximate_bias(batch_size)

    return self._sample_indices_with_bias(batch_size)


def _sample_indices_with_bias(self, batch_size: int) -> np.ndarray:
    valid_timestamps = self.timestamps[:self.size]
    age = self.current_time - valid_timestamps

    if self.linear_decay_step > 0:
        weights = np.maximum(
            self.min_weight,
            1.0 - age / self.abs_linear_decay_step,
        )
    else:
        weights = np.minimum(
            1.0,
            self.min_weight + age / self.abs_linear_decay_step,
        )

    if weights.sum() <= 0:
        weights = np.ones_like(weights, dtype=np.float32)

    probabilities = weights / weights.sum()
    return np.random.choice(self.size, size=batch_size, p=probabilities)
\end{lstlisting}

This exact sampler is faithful to the SWD definition, but it normalizes over all valid replay entries at each sampling call.

\paragraph{Approximate bucketed sampling.}
For CPU replay, we approximate SWD with fixed logical buckets. The sampler computes one midpoint weight per bucket, samples a bucket from the bucket-level distribution, and then samples uniformly within that bucket. GPU-parallel runs do not use this approximation.

\begin{lstlisting}[style=pythonappendix,caption={Approximate bucketed SWD sampling used to reduce wall-clock overhead.},label={lst:swd_approx_numpy}]
def _sample_indices_with_approximate_bias(self, batch_size: int) -> np.ndarray:
    bucket_size = max(
        (self.size + self.num_buckets - 1) // self.num_buckets,
        1,
    )

    logical_starts = np.arange(0, self.size, bucket_size)
    logical_ends = np.minimum(logical_starts + bucket_size, self.size)
    logical_midpoints = (logical_starts + logical_ends - 1) // 2

    if self.full:
        bucket_midpoints = (self.ptr + logical_midpoints) % self.max_size
    else:
        bucket_midpoints = logical_midpoints

    bucket_timestamps = self.timestamps[bucket_midpoints]
    bucket_ages = self.current_time - bucket_timestamps

    if self.linear_decay_step > 0:
        bucket_weights = np.maximum(
            self.min_weight,
            1.0 - bucket_ages / self.abs_linear_decay_step,
        )
    else:
        bucket_weights = np.minimum(
            1.0,
            self.min_weight + bucket_ages / self.abs_linear_decay_step,
        )

    if bucket_weights.sum() <= 0:
        bucket_weights = np.ones_like(bucket_weights, dtype=np.float32)

    bucket_probabilities = bucket_weights / bucket_weights.sum()
    sampled_buckets = np.random.choice(
        len(logical_starts),
        size=batch_size,
        p=bucket_probabilities,
    )

    sampled_starts = logical_starts[sampled_buckets]
    sampled_ends = logical_ends[sampled_buckets]
    offsets = (
        np.random.random(size=batch_size)
        * (sampled_ends - sampled_starts)
    ).astype(np.int64)
    logical_indices = sampled_starts + offsets

    if self.full:
        return (self.ptr + logical_indices) % self.max_size
    return logical_indices
\end{lstlisting}

The approximation is piecewise constant in replay age. We use 2000 buckets by default, making sampling cost depend mainly on the number of buckets and the minibatch size rather than replay capacity.

For circular buffers, logical indices are mapped through the current write pointer before indexing physical storage.

\subsection{Hyperparameters}
\label{app:hyperparameter}

Tables~\ref{tab:hyperparameter_profiles} and~\ref{tab:shared_hyperparameters} list the shared training settings. Unless stated otherwise, all variants within a benchmark family use the same environment interface, optimizer, replay buffer, evaluation protocol, and network size. The controlled changes are replay decay horizon, parameter projection normalization, and critic ensemble size. We use five seeds, $\{1,2,3,4,5\}$.

\begin{table}[H]
    \centering
    \caption{\textbf{Training profiles used across benchmark families.}
    CPU-scale domains use a single environment and CPU replay, while GPU-parallel domains use 1024 vectorized environments and GPU replay. Sim2Real retains the MJLab profile while doubling the training horizon to $100{,}001{,}792$ environment steps.}
    \label{tab:hyperparameter_profiles}
    \envtablesetup
    \setlength{\tabcolsep}{4pt}
    \begin{tabularx}{\textwidth}{@{}P{0.18\textwidth}CCCCCC@{}}
        \toprule
        \tablehead{Hyperparameter} & \tablehead{CPU-scale} & \tablehead{Playground} & \tablehead{MJLab} & \tablehead{ManiSkill} & \tablehead{IsaacLab} & \tablehead{Sim2Real} \\
        \midrule
        Benchmark families & DMC hard tasks, MuJoCo, MyoSuite, HumanoidBench & MuJoCo Playground & MJLab & ManiSkill & IsaacLab & Unitree-G1-Flat \\
        Number of environments & 1 & 1024 & 1024 & 1024 & 1024 & 1024 \\
        Total env. steps & $1.0{\times}10^6$ & $50{,}000{,}896$ & $50{,}000{,}896$ & $50{,}000{,}896$ & $50{,}000{,}896$ & $100{,}001{,}792$ \\
        Learning starts & $10{,}000$ & $100{,}000$ & $100{,}000$ & $100{,}000$ & $100{,}000$ & $100{,}000$ \\
        Replay buffer size & $1.0{\times}10^6$ & $1.0{\times}10^7$ & $1.0{\times}10^7$ & $1.0{\times}10^7$ & $1.0{\times}10^7$ & $1.0{\times}10^7$ \\
        Replay device & CPU & GPU & GPU & GPU & GPU & GPU \\
        Batch size & 512 & 2048 & 2048 & 2048 & 2048 & 2048 \\
        Grad. steps per interaction step & 1 & 2 & 2 & 2 & 2 & 2 \\
        Discount factor $\gamma$ & 0.99 & 0.97 & 0.99 & 0.90 & 0.99 & 0.99 \\
        $n$-step return & 1 & 1 & 3 & 1 & 3 & 3 \\
        SWD decay horizon & $80{,}000$ & $2{,}000$ & $2{,}000$ & $2{,}000$ & $2{,}000$ & $2{,}000$ \\
        Compute type & float32 & bfloat16 & bfloat16 & bfloat16 & bfloat16 & bfloat16 \\
        Evaluation episodes & 50 & 50 & 50 & 50 & 1024 & 50 \\
        Evaluation environments & 1 & 50 & 50 & 50 & 1024 & 50 \\
        \bottomrule
    \end{tabularx}
\end{table}

\begin{table}[H]
    \centering
    \caption{\textbf{Shared optimization and SAC hyperparameters.}
    These values are fixed unless a controlled ablation explicitly changes the corresponding component.}
    \label{tab:shared_hyperparameters}
    \envtablesetup
    \begin{tabularx}{0.90\textwidth}{@{}G{0.16\textwidth}LL@{}}
        \toprule
        & \tablehead{Hyperparameter} & \tablehead{Value} \\
        \midrule
        \tablegroup{3}{0.16\textwidth}{Optimization} & Actor learning rate & $3{\times}10^{-4}$ with cosine decay to $1.5{\times}10^{-4}$ \\
        & Critic learning rate & $3{\times}10^{-4}$ with cosine decay to $1.5{\times}10^{-4}$ \\
        & Temperature learning rate & $3{\times}10^{-4}$ with cosine decay to $1.5{\times}10^{-4}$ \\
        \midrule
        \tablegroup{2}{0.16\textwidth}{Entropy} & Initial entropy temperature & 0.01 \\
        & Target entropy & $\frac{1}{2}|\mathcal{A}|\log(2\pi e\cdot 0.15^2)$ \\
        \midrule
        \tablegroup{3}{0.16\textwidth}{Updates} & Target network update rate $\tau$ & 0.01 \\
        & Target update frequency & Every critic update \\
        & Policy update frequency & Every 2 critic updates \\
        \midrule
        \tablegroup{3}{0.16\textwidth}{Common} & Reward normalization & Enabled for all experiments \\
        & Action repeat & 1 \\
        & Linear-layer bias & Disabled \\
        \midrule
        \tablegroup{2}{0.16\textwidth}{Critic} & Critic ensemble size & 2 for FlashSAC, \method{} Norm ON, and \method{} Norm OFF; 1 for Single-Q variants \\
        & Critic output heads & 101 categorical heads with cross-entropy distributional critic loss \\
        \midrule
        \tablegroup{2}{0.16\textwidth}{Replay} & SWD minimum sampling weight & 0.1 \\
        & Uniform replay baseline & Implemented by setting the SWD decay horizon to 0 \\
        \bottomrule
    \end{tabularx}
\end{table}

For the normalization ablations, \textbf{Norm ON} applies parameter projection to both actor and critic after initialization and after each gradient update. \textbf{Norm OFF} disables this projection while keeping all other training settings fixed. The scale ablations in Section~\ref{sec:experiments} vary only the number of FlashSAC blocks in the actor and critic; the hidden dimensions, optimizer settings, replay settings, and environment budgets remain unchanged.

\subsection{Network Architecture}
\label{app:network}

All experiments use the same FlashSAC-style actor--critic backbone. The actor predicts a tanh-squashed Gaussian policy. The critic encodes the observation--action pair and predicts a distributional value head. Standard variants use two critics; Single-Q variants use one critic with the same per-critic architecture.

\begin{table}[H]
    \centering
    \caption{\textbf{Actor and critic architecture.}
    The number of residual blocks is 2 in the main experiments and is varied in the capacity ablations.}
    \label{tab:network_architecture}
    \envtablesetup
    \begin{tabularx}{0.96\textwidth}{@{}G{0.14\textwidth}LL@{}}
        \toprule
        & \tablehead{Component} & \tablehead{Architecture} \\
        \midrule
        \tablegroup{5}{0.14\textwidth}{Actor} & Input & Flattened actor observation; asymmetric environments use the actor observation stream for the policy. \\
        & Encoder & Batch-normalized input projection to 128 hidden units, followed by $B_\pi$ FlashSAC residual blocks and final RMSNorm. \\
        & Residual block & Linear $128\!\rightarrow\!512$, BatchNorm, ReLU, Linear $512\!\rightarrow\!128$, BatchNorm, ReLU, residual connection. \\
        & Output & Two orthogonally initialized linear heads predict action mean and log standard deviation. The policy samples from a tanh-squashed diagonal Gaussian and rescales actions to $[-1,1]$. \\
        & Log-standard-deviation range & $[-10,2]$ after tanh-based squashing. \\
        \midrule
        \tablegroup{4}{0.14\textwidth}{Critic} & Input & Concatenated critic observation and action; asymmetric environments use the critic observation stream for value learning. \\
        & Encoder & Batch-normalized input projection to 256 hidden units, followed by $B_Q$ FlashSAC residual blocks and final RMSNorm. \\
        & Residual block & Linear $256\!\rightarrow\!1024$, BatchNorm, ReLU, Linear $1024\!\rightarrow\!256$, BatchNorm, ReLU, residual connection. \\
        & Output & Orthogonally initialized linear projection from 256 hidden units to 101 categorical value logits. \\
        \midrule
        \tablegroup{4}{0.14\textwidth}{Variants} & Main setting & $B_\pi=B_Q=2$, with two critics for clipped double-Q learning. \\
        & Capacity ablation & $B_\pi=B_Q\in\{1,2,3\}$ depending on the experiment. \\
        & Single-Q variant & Same actor and per-critic network, but with one critic ensemble member and no clipped minimum over two critics. \\
        & Parameter projection normalization & When enabled, all linear kernels are projected by column-wise norm normalization. BatchNorm affine scale and bias are jointly normalized, and RMSNorm scales are normalized. \\
        \bottomrule
    \end{tabularx}
\end{table}

\section{Environment Details}
\label{app:environment}

This section lists the environments used in the experiments. The benchmark suite contains IsaacLab, MuJoCo Playground, MJLab, ManiSkill, Gym--MuJoCo, DMC hard tasks, HumanoidBench, and MyoSuite. It does not include Genesis environments.

\subsection{IsaacLab}

We use IsaacLab v2.1.0~\citep{mittal2025isaaclab}. The suite includes 12 gripper manipulation, dexterous manipulation, quadruped locomotion, and humanoid locomotion tasks. Scores are normalized using near-asymptotic FlashSAC references.

\begin{table}[H]
    \centering
    \caption{\textbf{IsaacLab environments.} Normalized scores correspond to near-asymptotic performance after extended training.}
    \label{tab:isaaclab_envs}
    \envtablesetup
    \begin{tabularx}{0.96\textwidth}{@{}Yccc@{}}
        \toprule
        \envhead{Task} & \envhead{Observation dim} & \envhead{Action dim} & \envhead{Normalize Score} \\
        \midrule
        Isaac-Repose-Cube-Shadow-Direct-v0 & 157 & 20 & 10000 \\
        Isaac-Repose-Cube-Allegro-Direct-v0 & 124 & 16 & 6000 \\
        Isaac-Velocity-Flat-G1-v0 & 123 & 37 & 40 \\
        Isaac-Velocity-Rough-G1-v0 & 310 & 37 & 40 \\
        Isaac-Velocity-Flat-H1-v0 & 69 & 19 & 40 \\
        Isaac-Velocity-Rough-H1-v0 & 256 & 19 & 40 \\
        Isaac-Lift-Cube-Franka-v0 & 36 & 8 & 160 \\
        Isaac-Open-Drawer-Franka-v0 & 31 & 8 & 100 \\
        Isaac-Velocity-Flat-Anymal-C-v0 & 48 & 12 & 30 \\
        Isaac-Velocity-Rough-Anymal-C-v0 & 235 & 12 & 30 \\
        Isaac-Velocity-Flat-Anymal-D-v0 & 48 & 12 & 30 \\
        Isaac-Velocity-Rough-Anymal-D-v0 & 235 & 12 & 30 \\
        \bottomrule
    \end{tabularx}
\end{table}

\subsection{MuJoCo Playground}

We evaluate four humanoid locomotion tasks from MuJoCo Playground v0.0.5~\citep{zakka2025mujoco}. Normalized scores are scaled to 40. Privileged critic observation dimensions are shown in parentheses when available.

\begin{table}[H]
    \centering
    \caption{\textbf{MuJoCo Playground environments.} We evaluate four humanoid locomotion tasks from MuJoCo Playground. Normalized scores are scaled to 40, corresponding to asymptotic performance achieved after extended training. If the environment supports asymmetric observation, the privileged observation size is given in parentheses.}
    \label{tab:playground_envs}
    \envtablesetup
    \begin{tabularx}{\textwidth}{@{}Yccc@{}}
        \toprule
        \envhead{Task} & \envhead{Observation dim} & \envhead{Action dim} & \envhead{Normalize Score} \\
        \midrule
        G1JoystickRoughTerrain & 103 (216) & 29 & 40 \\
        G1JoystickFlatTerrain & 103 (216) & 29 & 40 \\
        T1JoystickRoughTerrain & 85 (180) & 23 & 40 \\
        T1JoystickFlatTerrain & 85 (180) & 23 & 40 \\
        \bottomrule
    \end{tabularx}
\end{table}

\subsection{MJLab}
\label{app:env-mjlab}

We use MJLab v1.2.0~\citep{zakka2026mjlab}, an Isaac Lab-style robot-learning framework powered by the MuJoCo-Warp GPU simulator. We evaluate eight GPU-parallel tasks: four Unitree velocity-tracking locomotion tasks and four Yam manipulation tasks. Normalized scores are scaled to 100 for the locomotion tasks and 20 for the Yam manipulation tasks. Privileged critic observation dimensions are shown in parentheses when available.

\begin{table}[H]
    \centering
    \caption{\textbf{MJLab environments.} We evaluate Unitree velocity-tracking and Yam manipulation tasks from MJLab. Observation dimensions report the actor input, with the effective critic input shown in parentheses. Normalized scores are scaled to 100 for locomotion tasks and 20 for manipulation tasks.}
    \label{tab:mjlab_envs}
    \envtablesetup
    \begin{tabularx}{\textwidth}{@{}Yccc@{}}
        \toprule
        \envhead{Task} & \envhead{Observation dim} & \envhead{Action dim} & \envhead{Normalize Score} \\
        \midrule
        Mjlab-Velocity-Flat-Unitree-G1 & 99 (210) & 29 & 100 \\
        Mjlab-Velocity-Rough-Unitree-G1 & 286 (584) & 29 & 100 \\
        Mjlab-Velocity-Flat-Unitree-Go1 & 48 (120) & 12 & 100 \\
        Mjlab-Velocity-Rough-Unitree-Go1 & 235 (494) & 12 & 100 \\
        Mjlab-Lift-Cube-Yam & 29 (58) & 7 & 20 \\
        Mjlab-Lift-Cube-Yam-Depth & 26 (55) & 7 & 20 \\
        Mjlab-Lift-Cube-Yam-Rgb & 26 (55) & 7 & 20 \\
        Mjlab-Multi-Cube-Seg-Yam & 26 (52) & 7 & 20 \\
        \bottomrule
    \end{tabularx}
\end{table}

\subsection{ManiSkill}

We use ManiSkill~\citep{taomaniskill3}, a large-scale benchmark built on the SAPIEN physics engine. We evaluate six rigid-body manipulation tasks with a gripper-based robotic arm. For reproducibility, we use the environment snapshot corresponding to commit hash \texttt{aad75f2}. Normalized scores correspond to the maximum success rate, where 100 denotes 100\% success.

\begin{table}[H]
    \centering
    \caption{\textbf{ManiSkill environments.} We evaluate 6 ManiSkill gripper-based manipulation environments. Normalized scores correspond to the maximum success rate, where 100 denotes a 100\% success rate.}
    \label{tab:maniskill_envs}
    \envtablesetup
    \begin{tabularx}{0.80\textwidth}{@{}Yccc@{}}
        \toprule
        \envhead{Task} & \envhead{Observation dim} & \envhead{Action dim} & \envhead{Normalize Score} \\
        \midrule
        PickSingleYCB-v1 & 45 & 8 & 100 \\
        PegInsertionSide-v1 & 43 & 8 & 100 \\
        LiftPegUpright-v1 & 32 & 8 & 100 \\
        PokeCube-v1 & 54 & 8 & 100 \\
        PullCube-v1 & 35 & 8 & 100 \\
        RollBall-v1 & 44 & 8 & 100 \\
        \bottomrule
    \end{tabularx}
\end{table}

\subsection{Gym--MuJoCo}

We evaluate five Gym--MuJoCo v4 locomotion tasks~\citep{todorov2012mujoco}. Scores are normalized using random-policy lower references and TD7 5M-step upper references~\citep{fujimoto2023td7}.

\begin{table}[H]
    \centering
    \caption{\textbf{Gym--MuJoCo environments.} Normalized scores are computed relative to random-policy and TD7 5M-step reference scores.}
    \label{tab:mujoco_envs}
    \envtablesetup
    \begin{tabularx}{\textwidth}{@{}Ycccc@{}}
        \toprule
        \envhead{Task} & \envhead{Observation dim} & \envhead{Action dim} & \envhead{Random Score} & \envhead{Normalize Score} \\
        \midrule
        HalfCheetah-v4 & 17 & 6 & -289.415 & 18165 \\
        Hopper-v4 & 11 & 3 & 18.791 & 4075 \\
        Walker2d-v4 & 17 & 6 & 2.791 & 7397 \\
        Ant-v4 & 27 & 8 & -70.288 & 10133 \\
        Humanoid-v4 & 376 & 17 & 104.361 & 8584 \\
        \bottomrule
    \end{tabularx}
\end{table}

\subsection{DeepMind Control Suite Hard Tasks}

We report only DMC hard tasks: humanoid and dog domains from the DeepMind Control Suite~\citep{tassa2018deepmind}. Low-dimensional tasks such as Cartpole and Pendulum are not included. Normalized scores use the DMC maximum of 1000.

\begin{table}[H]
    \centering
    \caption{\textbf{DMC hard environments.} We report only humanoid and dog tasks, with normalized scores set to the DMC maximum.}
    \label{tab:dmc_envs}
    \envtablesetup
    \begin{tabularx}{0.80\textwidth}{@{}Yccc@{}}
        \toprule
        \envhead{Task} & \envhead{Observation dim} & \envhead{Action dim} & \envhead{Normalize Score} \\
        \midrule
        humanoid-stand & 67 & 21 & 1000 \\
        humanoid-walk & 67 & 21 & 1000 \\
        humanoid-run & 67 & 21 & 1000 \\
        dog-stand & 223 & 38 & 1000 \\
        dog-walk & 223 & 38 & 1000 \\
        dog-run & 223 & 38 & 1000 \\
        dog-trot & 223 & 38 & 1000 \\
        \bottomrule
    \end{tabularx}
\end{table}

\subsection{HumanoidBench}

We evaluate HumanoidBench~\citep{sferrazza2024humanoidbench}, a high-dimensional whole-body-control benchmark based on the Unitree H1 humanoid robot. In our experiments, HumanoidBench is installed from the compatibility fork \texttt{https://github.com/wzhhasadream/humanoid-bench-compatible.git}. We evaluate 14 locomotion and whole-body-control tasks without hand control. Normalized scores follow the benchmark success thresholds.

\begin{table}[H]
    \centering
    \caption{\textbf{HumanoidBench environments.} Scores are normalized using benchmark task-success thresholds.}
    \label{tab:humanoidbench_envs}
    \envtablesetup
    \begin{tabularx}{0.92\textwidth}{@{}Ycccc@{}}
        \toprule
        \envhead{Task} & \envhead{Observation dim} & \envhead{Action dim} & \envhead{Random Score} & \envhead{Normalize Score} \\
        \midrule
        h1-walk-v0 & 51 & 19 & 2.38 & 700 \\
        h1-stand-v0 & 51 & 19 & 10.55 & 800 \\
        h1-run-v0 & 51 & 19 & 2.02 & 700 \\
        h1-reach-v0 & 57 & 19 & 260.30 & 12000 \\
        h1-maze-v0 & 51 & 19 & 106.44 & 1200 \\
        h1-hurdle-v0 & 51 & 19 & 2.21 & 700 \\
        h1-crawl-v0 & 51 & 19 & 272.66 & 700 \\
        h1-sit\_simple-v0 & 51 & 19 & 9.40 & 750 \\
        h1-sit\_hard-v0 & 64 & 19 & 2.45 & 750 \\
        h1-balance\_simple-v0 & 64 & 19 & 9.40 & 800 \\
        h1-balance\_hard-v0 & 77 & 19 & 9.04 & 800 \\
        h1-stair-v0 & 51 & 19 & 3.11 & 700 \\
        h1-slide-v0 & 51 & 19 & 3.19 & 700 \\
        h1-pole-v0 & 51 & 19 & 20.09 & 700 \\
        \bottomrule
    \end{tabularx}
\end{table}

\subsection{MyoSuite}

We evaluate 10 MyoSuite dexterous manipulation tasks~\citep{caggiano2022myosuite}. Tasks marked \texttt{hard} use randomized goals. Normalized scores correspond to success rate, with 100 denoting 100\% success.

\begin{table}[H]
    \centering
    \caption{\textbf{MyoSuite environments.} Normalized scores correspond to 100\% success.}
    \label{tab:myosuite_envs}
    \envtablesetup
    \begin{tabularx}{0.80\textwidth}{@{}Yccc@{}}
        \toprule
        \envhead{Task} & \envhead{Observation dim} & \envhead{Action dim} & \envhead{Normalize Score} \\
        \midrule
        myo-reach & 115 & 39 & 100 \\
        myo-reach-hard & 115 & 39 & 100 \\
        myo-pose & 108 & 39 & 100 \\
        myo-pose-hard & 108 & 39 & 100 \\
        myo-obj-hold & 91 & 39 & 100 \\
        myo-obj-hold-hard & 91 & 39 & 100 \\
        myo-key-turn & 93 & 39 & 100 \\
        myo-key-turn-hard & 93 & 39 & 100 \\
        myo-pen-twirl & 83 & 39 & 100 \\
        myo-pen-twirl-hard & 83 & 39 & 100 \\
        \bottomrule
    \end{tabularx}
\end{table}

\subsection{Sim2Real}
\label{app:env-sim2real}

For sim-to-real locomotion, we use \texttt{Unitree-G1-Flat} from Unitree Robotics' \href{https://github.com/unitreerobotics/unitree_rl_mjlab}{\texttt{unitree\_rl\_mjlab}} implementation. The task exposes asymmetric observations: the policy receives a 98-dimensional actor observation, while the critic receives a 211-dimensional privileged observation. The 29-dimensional action controls the Unitree G1 joints.

\begin{table}[H]
    \centering
    \caption{\textbf{Sim2Real environment.} The \texttt{Unitree-G1-Flat} locomotion task uses asymmetric actor and critic observations for sim-to-real policy training.}
    \label{tab:sim2real_env}
    \envtablesetup
    \begin{tabularx}{0.76\textwidth}{@{}Yccc@{}}
        \toprule
        \envhead{Task} & \envhead{Actor obs. dim} & \envhead{Critic obs. dim} & \envhead{Action dim} \\
        \midrule
        Unitree-G1-Flat & 98 & 211 & 29 \\
        \bottomrule
    \end{tabularx}
\end{table}

\clearpage
\section{Full Results}
\label{app:results}

\subsection{Gym--MuJoCo}
\label{app:results-mujoco}

\begin{figure}[H]
    \centering
    \vspace{-0.15cm}
    \includegraphics[width=\linewidth]{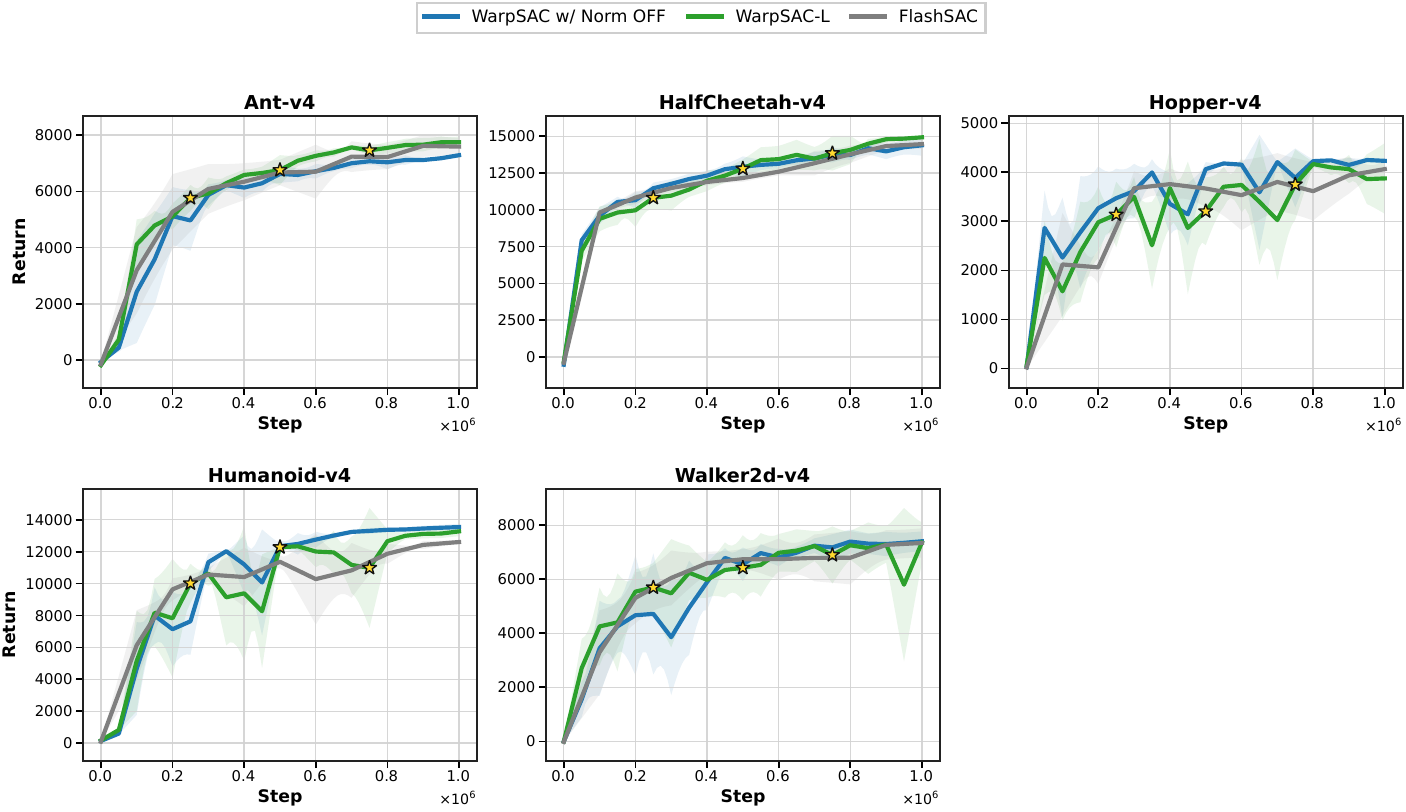}
    \vspace{-0.2cm}
    \caption{\textbf{Full Gym--MuJoCo learning curves.}
    Curves report mean performance with standard deviation across five seeds for all evaluated Gym--MuJoCo tasks.}
    \label{fig:full_mujoco}
    \vspace{-0.3cm}
\end{figure}

Figure~\ref{fig:full_mujoco} reports the complete Gym--MuJoCo learning curves. These tasks provide a standard CPU-scale locomotion benchmark for comparing final performance and sample efficiency under single-environment replay.

\Needspace{0.75\textheight}
\subsection{DMC}
\label{app:results-dmc}

\begin{figure}[H]
    \centering
    \vspace{-0.15cm}
    \includegraphics[width=\linewidth]{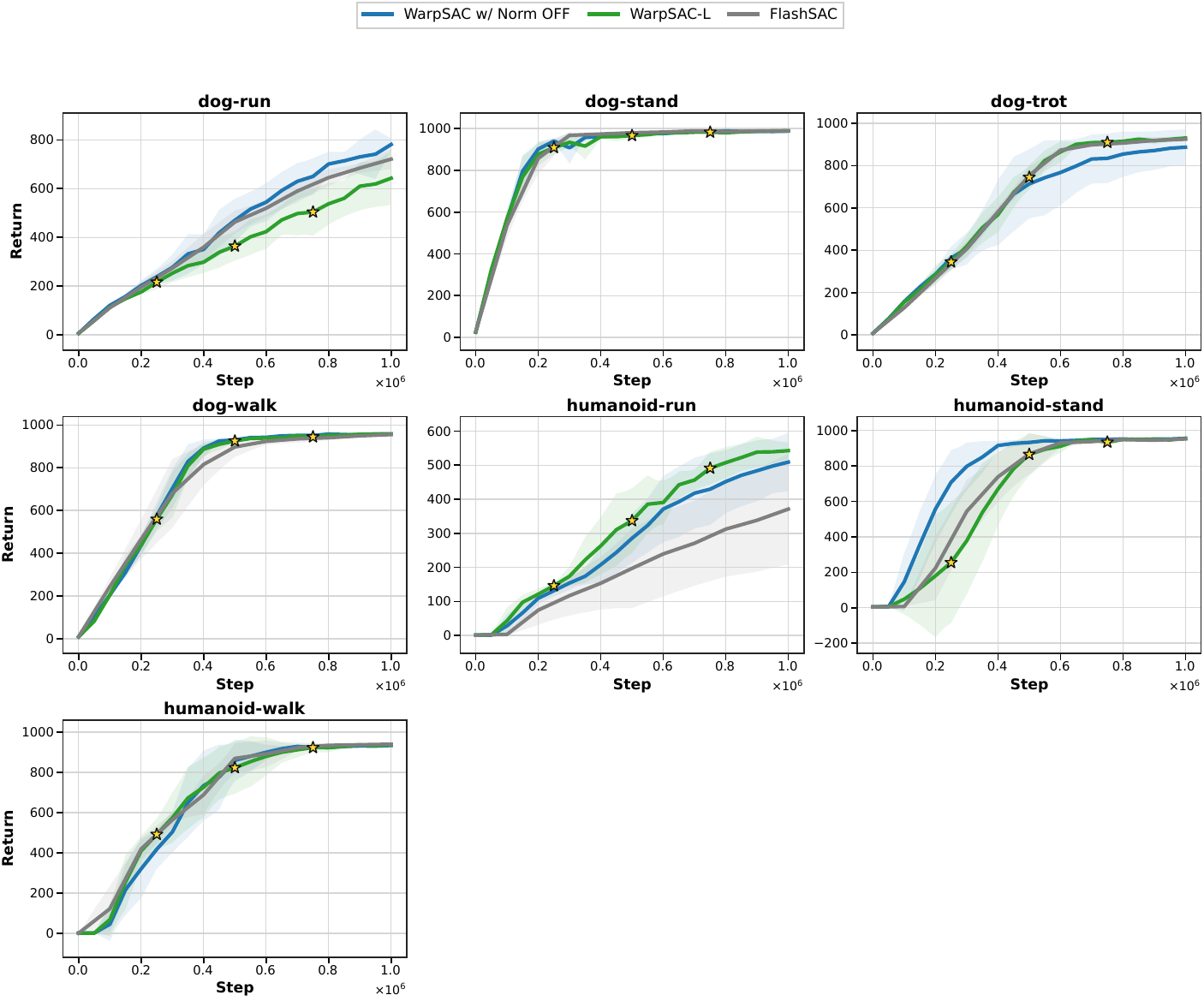}
    \vspace{-0.2cm}
    \caption{\textbf{Full DMC hard-task learning curves.}
    Curves report mean performance with standard deviation across seeds on humanoid and dog tasks from the DeepMind Control Suite.}
    \label{fig:full_dmc}
    \vspace{-0.3cm}
\end{figure}

Figure~\ref{fig:full_dmc} reports the full DMC hard-task results on humanoid and dog domains. These tasks test whether replay weighting remains useful in CPU-scale control settings with diverse dynamics and reward scales.

\Needspace{0.85\textheight}
\subsection{MyoSuite}
\label{app:results-myosuite}

\begin{figure}[H]
    \centering
    \vspace{-0.15cm}
    \includegraphics[width=0.80\linewidth]{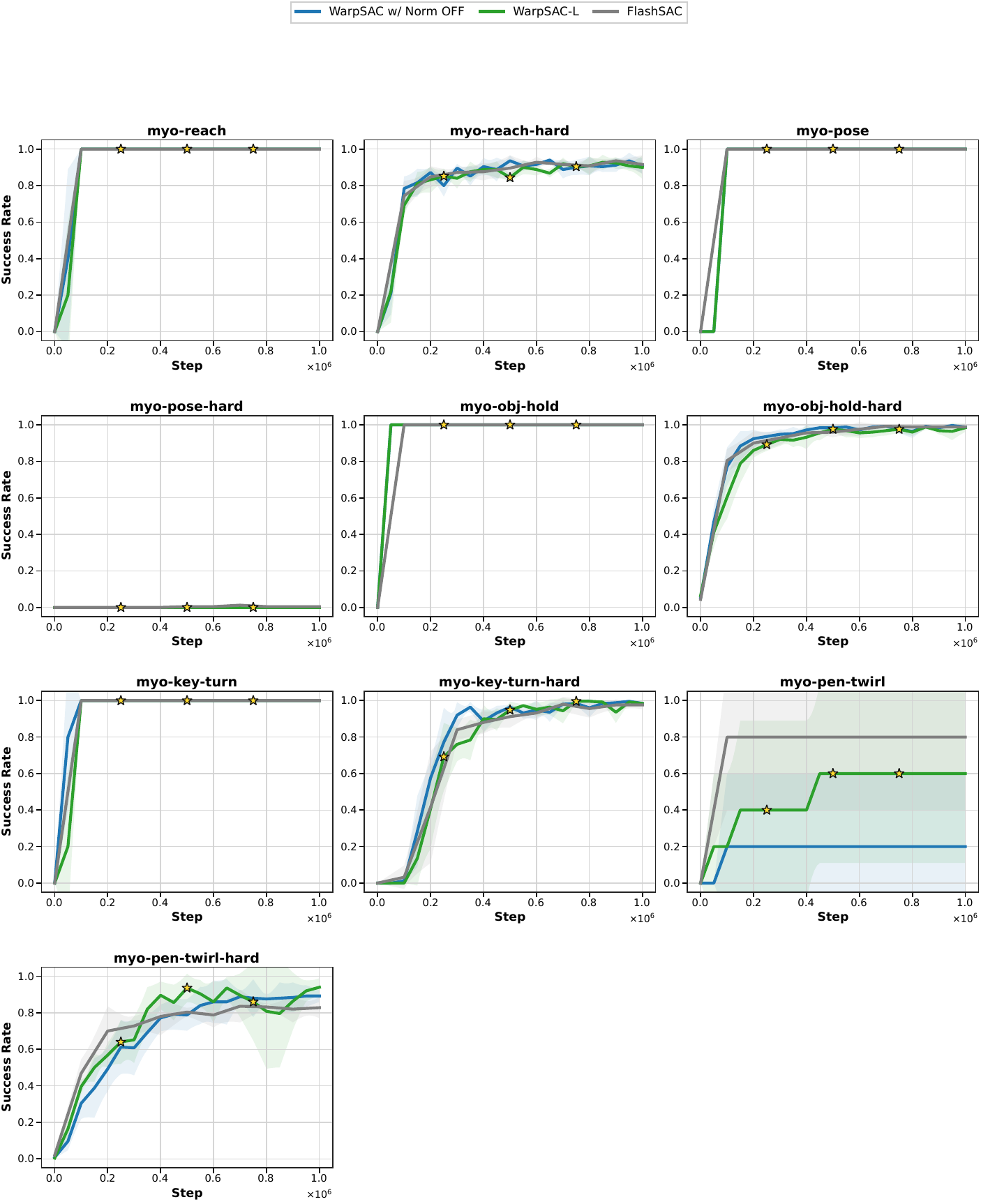}
    \vspace{-0.2cm}
    \caption{\textbf{Full MyoSuite learning curves.}
    Curves report mean success-oriented performance with standard deviation across five seeds for the evaluated MyoSuite manipulation tasks.}
    \label{fig:full_myosuite}
    \vspace{-0.3cm}
\end{figure}

Figure~\ref{fig:full_myosuite} reports the dexterous manipulation results on MyoSuite. These environments use success-rate-style returns, so the curves emphasize how quickly each method reaches reliable task completion.

\Needspace{0.80\textheight}
\subsection{HumanoidBench}
\label{app:results-humanoidbench}

\begin{figure}[H]
    \centering
    \vspace{-0.15cm}
    \includegraphics[width=\linewidth]{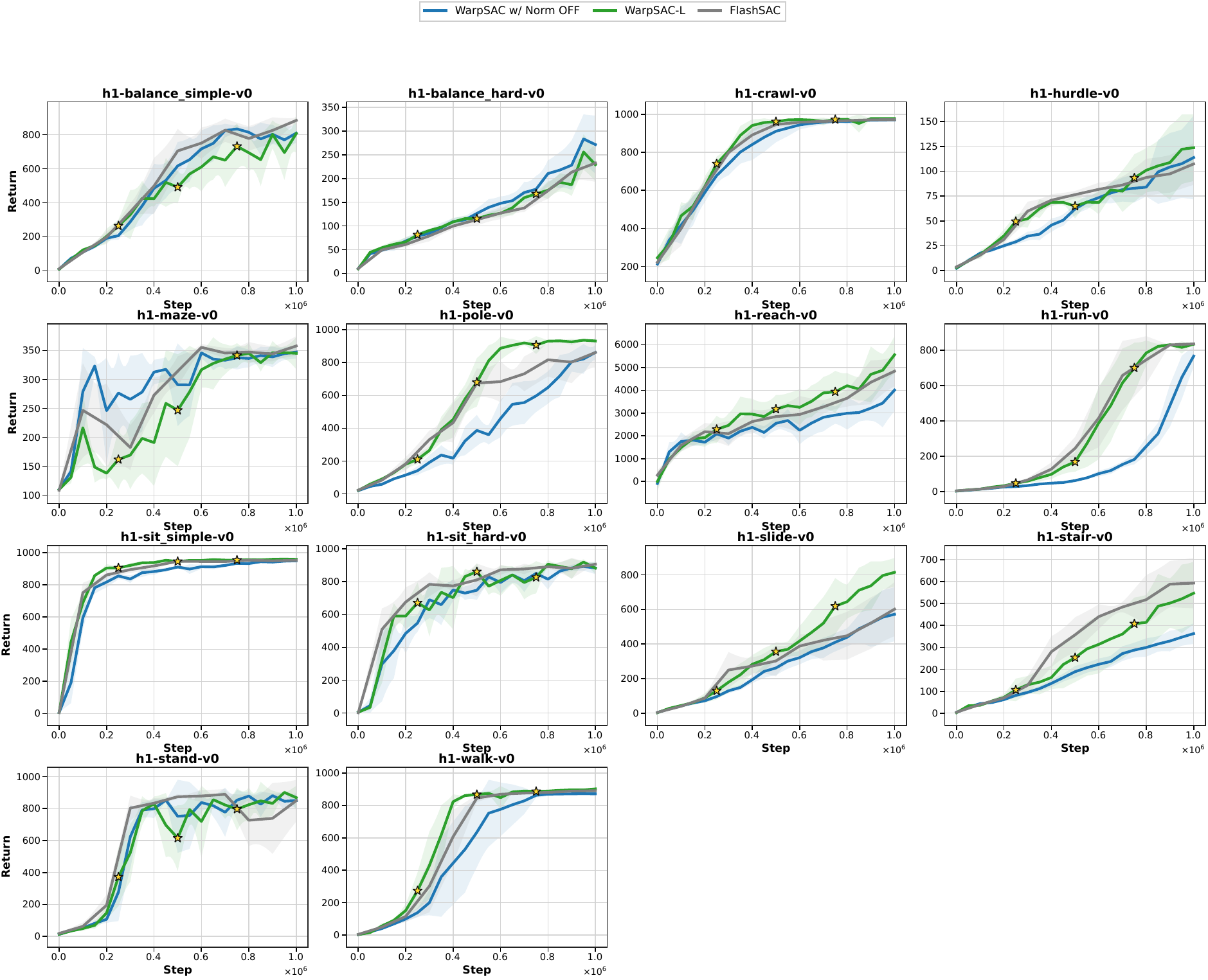}
    \vspace{-0.2cm}
    \caption{\textbf{Full HumanoidBench learning curves.}
    Curves report mean performance with standard deviation across five seeds for the evaluated H1 whole-body-control tasks.}
    \label{fig:full_humanoidbench}
    \vspace{-0.3cm}
\end{figure}

Figure~\ref{fig:full_humanoidbench} gives the full HumanoidBench results. These tasks are high-dimensional whole-body-control problems and provide a stress test for stability under sparse or delayed progress.

\Needspace{0.80\textheight}
\subsection{MuJoCo Playground}
\label{app:results-playground}

\begin{figure}[H]
    \centering
    \vspace{-0.15cm}
    \includegraphics[width=\linewidth]{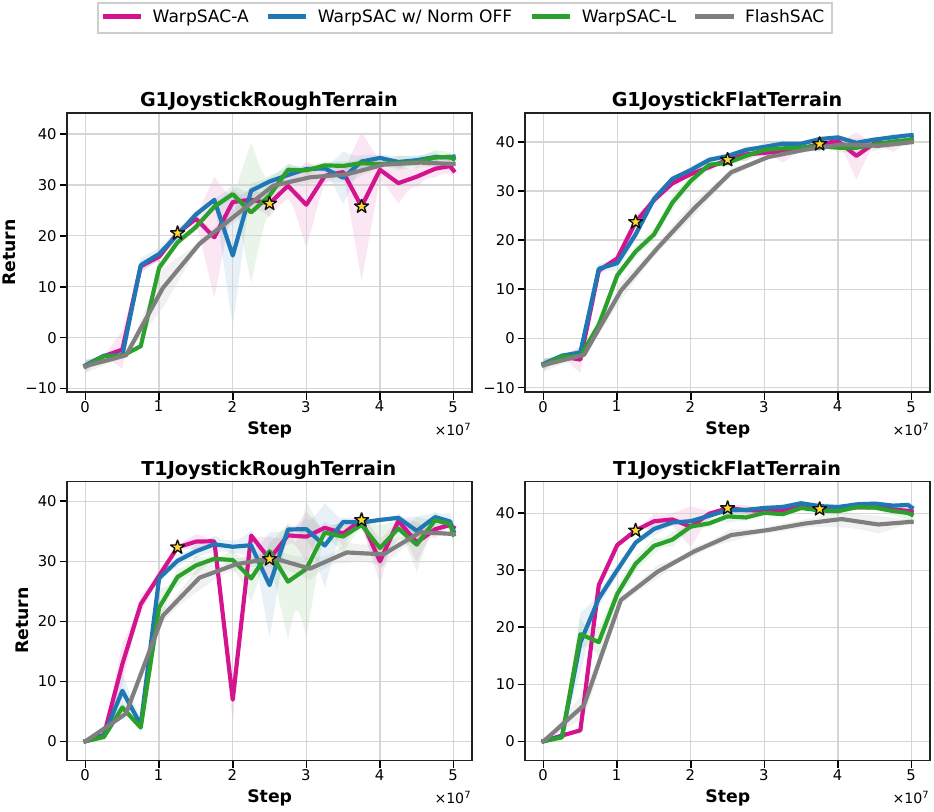}
    \vspace{-0.2cm}
    \caption{\textbf{Full MuJoCo Playground learning curves.}
    Curves report mean performance with standard deviation across five seeds for the evaluated GPU-parallel humanoid locomotion tasks.}
    \label{fig:full_playground}
    \vspace{-0.3cm}
\end{figure}

Figure~\ref{fig:full_playground} reports the GPU-parallel MuJoCo Playground results. The large number of vectorized environments shifts the comparison toward how effectively each method exploits broad replay coverage.

\Needspace{0.85\textheight}
\subsection{IsaacLab}
\label{app:results-isaaclab}

\begin{figure}[H]
    \centering
    \vspace{-0.15cm}
    \includegraphics[width=0.80\linewidth]{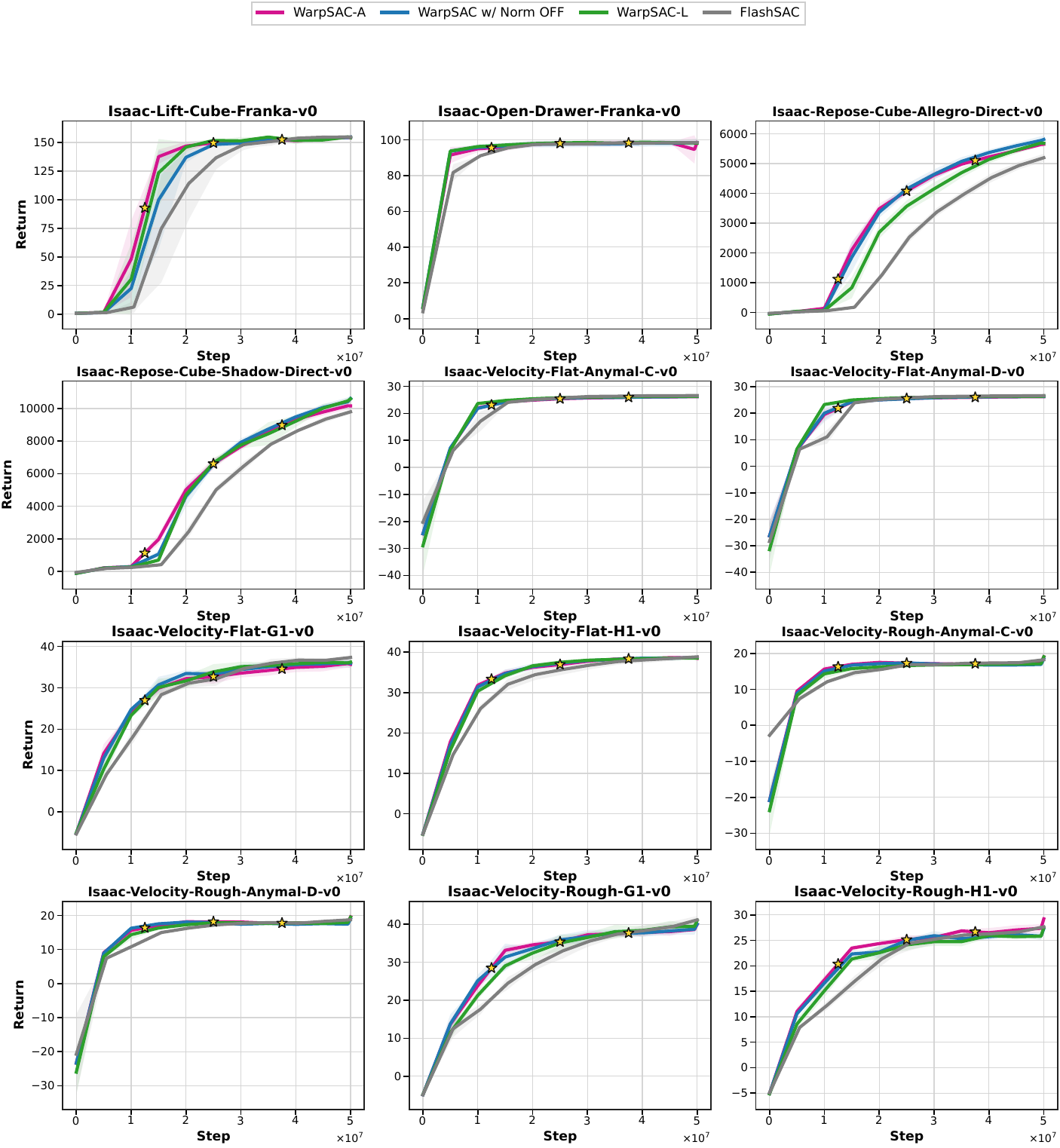}
    \vspace{-0.2cm}
    \caption{\textbf{Full IsaacLab learning curves.}
    Curves report mean performance with standard deviation across five seeds for the evaluated IsaacLab manipulation and locomotion tasks.}
    \label{fig:full_isaaclab}
    \vspace{-0.3cm}
\end{figure}

Figure~\ref{fig:full_isaaclab} reports the full IsaacLab results across manipulation and locomotion tasks. This domain evaluates performance under highly parallel simulation with diverse robot embodiments and reward scales.

\Needspace{0.80\textheight}
\subsection{MJLab}
\label{app:results-mjlab}

\begin{figure}[H]
    \centering
    \vspace{-0.15cm}
    \includegraphics[width=\linewidth]{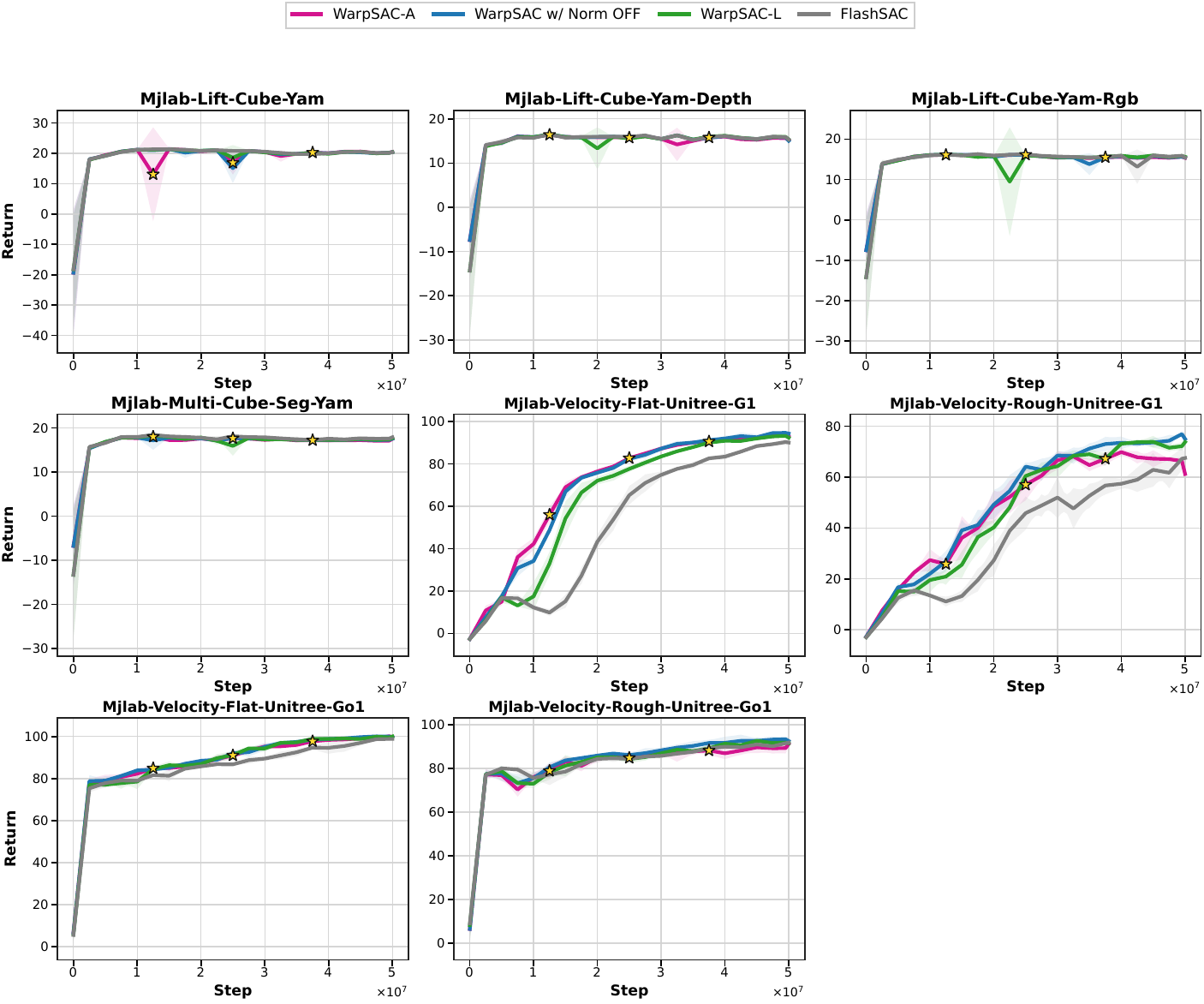}
    \vspace{-0.2cm}
    \caption{\textbf{Full MJLab learning curves.}
    Curves report mean episodic return with standard deviation across five seeds for Unitree locomotion and Yam manipulation tasks.}
    \label{fig:full_mjlab}
    \vspace{-0.3cm}
\end{figure}

Figure~\ref{fig:full_mjlab} reports the complete MJLab results, covering Unitree G1 and Go1 velocity-tracking tasks on flat and rough terrain as well as Yam manipulation tasks with state, depth, RGB, and segmentation observations. These tasks provide an additional GPU-parallel backend spanning locomotion and manipulation, allowing us to test whether the observed data-regime effects extend beyond IsaacLab and MuJoCo Playground.

\Needspace{0.80\textheight}
\subsection{ManiSkill}
\label{app:results-maniskill}

\begin{figure}[H]
    \centering
    \vspace{-0.15cm}
    \includegraphics[width=\linewidth]{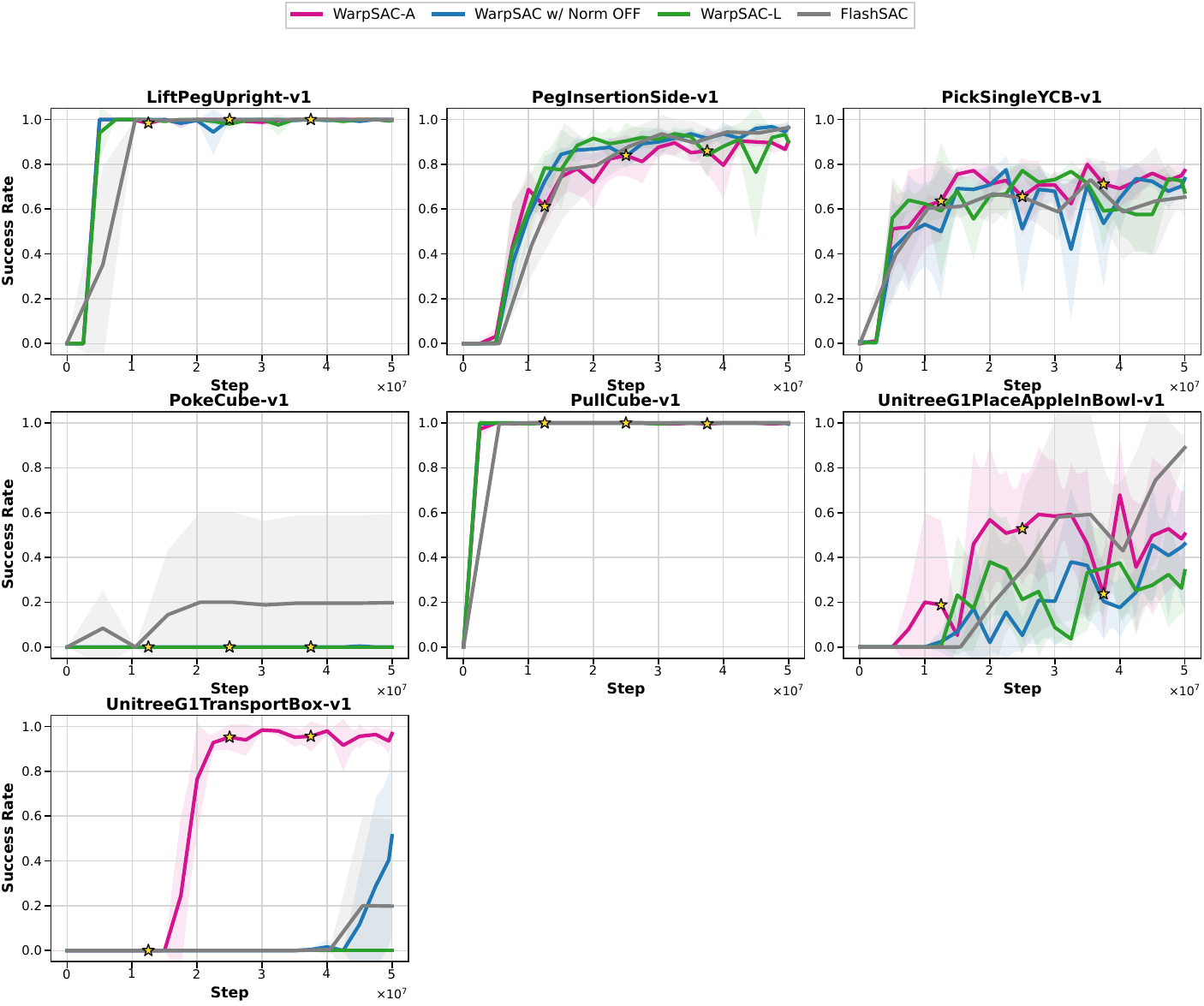}
    \vspace{-0.2cm}
    \caption{\textbf{Full ManiSkill learning curves.}
    Curves report mean success-oriented performance with standard deviation across five seeds for the evaluated ManiSkill manipulation tasks.}
    \label{fig:full_maniskill}
    \vspace{-0.3cm}
\end{figure}

Figure~\ref{fig:full_maniskill} reports the full ManiSkill results. These GPU-parallel manipulation tasks test whether the methods remain effective when success-rate objectives are collected from many simultaneous environments.

\clearpage
\subsection{Ablation Learning Curves}
\label{app:results-ablation}

\begin{figure}[H]
    \centering
    \vspace{-0.15cm}
    \includegraphics[width=\linewidth]{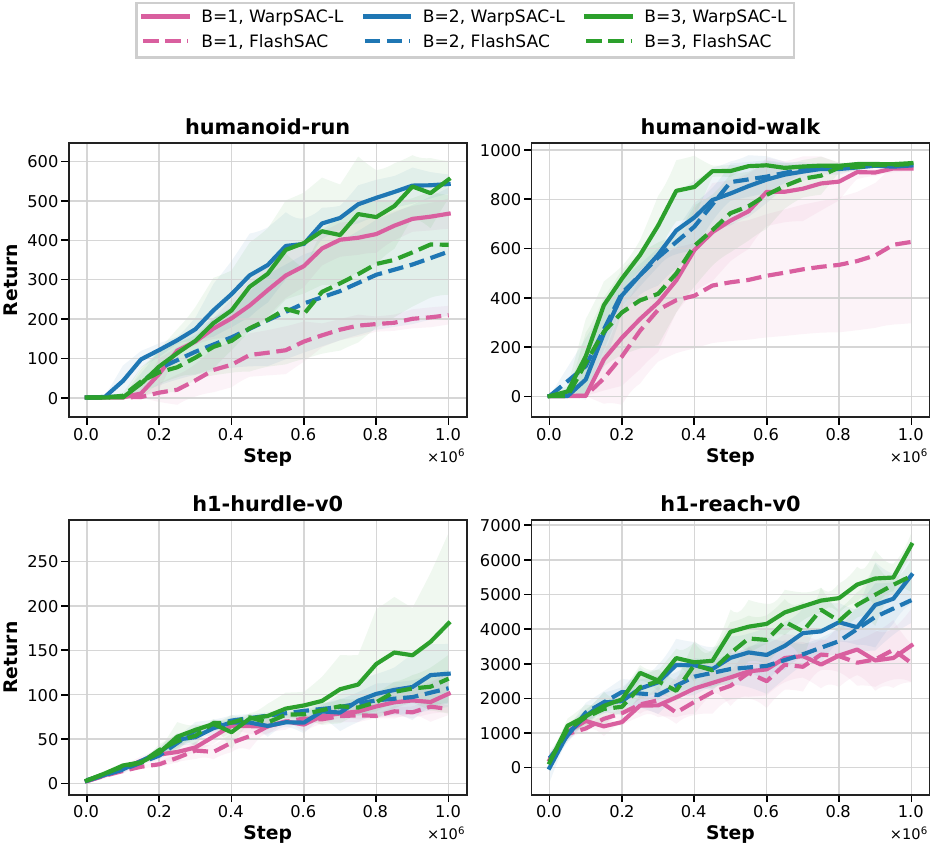}
    \vspace{-0.2cm}
    \caption{\textbf{Capacity and SWD ablation learning curves.}
    Curves compare SWD and uniform replay using one, two, and three residual blocks on humanoid-run, humanoid-walk, h1-hurdle-v0, and h1-reach-v0. Shaded regions show standard deviation across five seeds.}
    \label{fig:full_scale_ablation_curves}
    \vspace{-0.3cm}
\end{figure}

\begin{figure}[H]
    \centering
    \vspace{-0.15cm}
    \includegraphics[width=\linewidth]{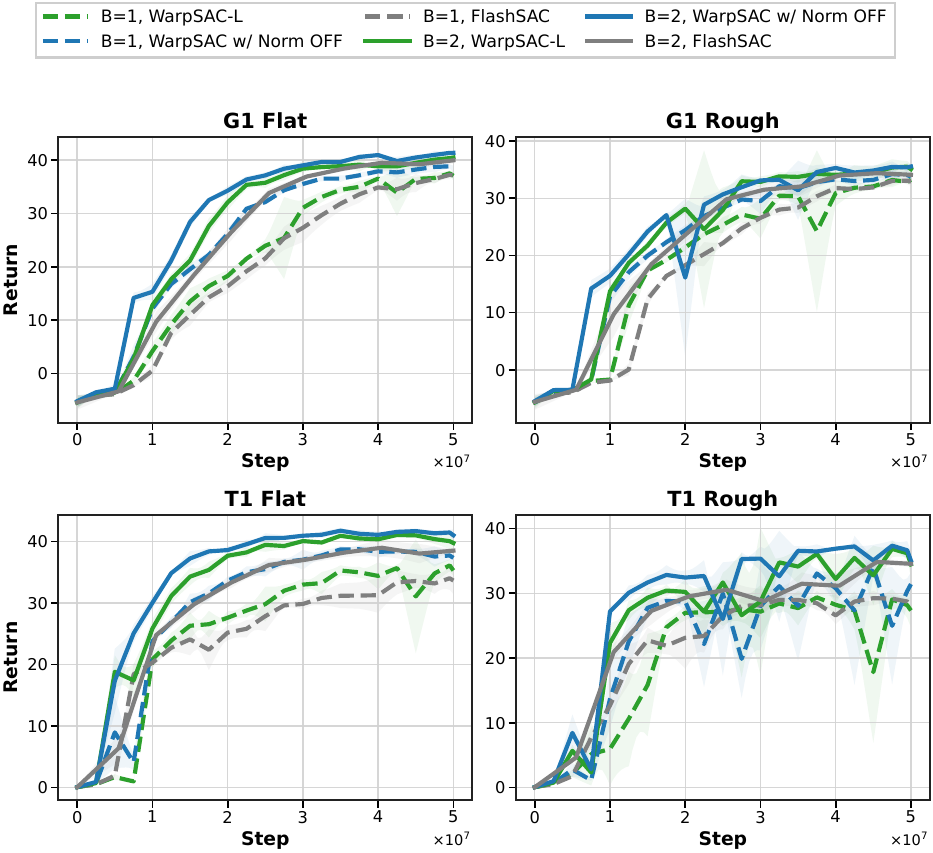}
    \vspace{-0.2cm}
    \caption{\textbf{MuJoCo Playground normalization and capacity ablation learning curves.}
    Curves compare Norm ON, Norm OFF, and No SWD using one and two residual blocks on the four evaluated Playground tasks. Shaded regions show standard deviation across five seeds.}
    \label{fig:full_playground_ablation_curves}
    \vspace{-0.3cm}
\end{figure}

The endpoint and AUC bar charts in the main paper summarize the controlled ablations. Figures~\ref{fig:full_scale_ablation_curves} and~\ref{fig:full_playground_ablation_curves} provide the corresponding learning curves, exposing how the capacity, replay-weighting, and normalization choices affect learning speed throughout training.

\end{document}